\documentclass[pdflatex]{sn-jnl}

\usepackage{graphicx}%
\usepackage{multirow}%
\usepackage{amsmath,amssymb,amsfonts}%
\usepackage{amsthm}%
\usepackage{mathrsfs}%
\usepackage[title]{appendix}%
\usepackage{xcolor}%
\usepackage{textcomp}%
\usepackage{manyfoot}%
\usepackage{booktabs}%
\usepackage{algorithm}%
\usepackage{algorithmicx}%
\usepackage{algpseudocode}%
\usepackage{listings}%
\usepackage{hyperref}
\usepackage[authoryear]{natbib}
\setcitestyle{aysep={}} 

\theoremstyle{thmstyleone}%
\theoremstyle{thmstyletwo}%
\newtheorem{remark}{Remark}%

\theoremstyle{thmstylethree}%
\newtheorem{definition}{Definition}%

\begin{document}

\title[ffstruc2vec: Flat, Flexible and Scalable]{ffstruc2vec: Flat, Flexible and Scalable Learning of Node Representations from Structural Identities}


\author*[1,2]{\fnm{Mario} \sur{Heidrich} \href{https://orcid.org/0009-0009-4875-1325}{\textsuperscript{ORCID}}}\email{mheidrich@alu.ucam.edu}
\author[2]{\fnm{Jeffrey} \sur{Heidemann} \href{https://orcid.org/0009-0008-8467-1026}{\textsuperscript{ORCID}}}\email{jeffrey.heidemann@fom.de}
\author[2]{\fnm{Rüdiger} \sur{Buchkremer} \href{https://orcid.org/0000-0002-4130-9253}{\textsuperscript{ORCID}}}\email{ruediger.buchkremer@fom.de}
\author[1]{\fnm{Gonzalo Wandosell} \sur{Fernández de Bobadilla}\href{https://orcid.org/0000-0002-0617-5480}{\textsuperscript{ORCID}}}\email{gwandosell@ucam.edu}

\affil[1]{\orgname{Universidad Católica San Antonio de Murcia (UCAM)}, \orgaddress{\street{Campus de los Jerónimos, nº 135}, \city{Murcia}, \postcode{30107}, \country{Spain}}}
\affil[2]{\orgdiv{Institute of IT Management and Digitization Research (IFID)}, \orgname{FOM University of Applied Sciences in Economics and Management}, \orgaddress{\street{Toulouser Allee 53}, \city{Düsseldorf}, \postcode{40476}, \state{North Rhine‑Westphalia}, \country{Germany}}}


\abstract{Node embedding refers to techniques that generate low-dimensional vector representations of nodes in a graph while preserving specific properties of the nodes. A key challenge in the field is developing scalable methods that can preserve structural properties suitable for the required types of structural patterns of a given downstream application task. While most existing methods focus on preserving node proximity, those that do preserve structural properties often lack the flexibility to preserve various types of structural patterns required by downstream application tasks.

This paper introduces \textit{ffstruc2vec}, a scalable deep-learning framework for learning node embedding vectors that preserve structural identities. Its flat, efficient architecture allows high flexibility in capturing diverse types of structural patterns, enabling broad adaptability to various downstream application tasks. The proposed framework significantly outperforms existing approaches across diverse unsupervised and supervised tasks in practical applications. Moreover, ffstruc2vec enables explainability by quantifying how individual structural patterns influence task outcomes, providing actionable interpretation. To our knowledge, no existing framework combines this level of flexibility, scalability, and structural interpretability, underscoring its unique capabilities.}




\maketitle

%
%

\section{Introduction}
\label{sec:Introduction}

Node embedding is the process of generating low-dimensional vector representations of nodes in a graph, preserving task-relevant properties of nodes, including but not limited to structure, proximity, or attributes. These embeddings can be leveraged in various downstream tasks, including node classification, link prediction, clustering, exploratory data analysis, and network visualization. The method has found broad application across diverse domains, such as fraud detection in financial networks \citep{vanBelle.2023}, friendship recommendation and bot detection in social networks \citep{Saxena.2022, Dehghan.2023}, knowledge discovery in knowledge graphs \citep{Egami.2023, Liu.2023}, analysis of biological networks \citep{Jiang.2021, Pasquier.2023}, and fake review detection on online platforms \citep{Zaki.2024}.

A key challenge in Node Embedding is developing a scalable method for preserving the structural properties of nodes suitable for the required structural patterns of a downstream application task. The type of structural patterns in which a node is embedded within the graph can vary depending on the role or function of the node in a specific application task. For instance, fraudulent activities such as money laundering can be embedded in particular money flow patterns among illicit entities, resulting in characteristic structural patterns within the financial transaction network, such as suspicious cyclic transaction chains \citep{Granados.2022}. These structural patterns differ significantly from those observed in social networks, where specific roles such as bridge and core nodes define the network's connectivity and influence \citep{Huang.2014}. As Node Embedding methods cannot preserve all types of structural patterns simultaneously, they must align with the requirements of a specific application task when defining types of structural identities.

ffstruc2vec is a flexible framework capable of integrating diverse types of structural patterns, including complex ones. If specific structural patterns, such as suspicious cyclic transaction chains in fraud detection, are known to be relevant, domain-specific knowledge can be incorporated manually. At the same time, ffstruc2vec can automatically adapt to task-specific requirements by optimizing corresponding weights in a supervised manner. The optimized weights provide insights into the relevance of specific structural patterns for a given application. The resulting interpretability and explainability are particularly crucial in highly regulated domains such as fraud detection. To the best of our knowledge, these characteristics make ffstruc2vec unique among existing node embedding frameworks.

A detailed overview of node embedding frameworks, their limitations, and relevant references can be found in Section~\ref{sec:Related_Work}. Most node embedding frameworks primarily focus on preserving the proximity of nodes. Frameworks that instead aim to capture structural identities often struggle to adapt effectively to the specific types of structural identities required by downstream application tasks. For instance, struc2vec, which employs a multilayer graph to encode structural similarities, is limited in its ability to capture diverse types of structural identities. The ffstruc2vec approach addresses this issue more effectively by implementing a flexible and flattened similarity graph encoding structural similarities of the nodes. This approach involves assessing structural similarity between nodes by analyzing their respective structural properties and those of their surrounding neighborhoods. This evaluation incorporates various structural properties, including centrality measures, clustering coefficients, attached graphlets, and anonymous walk-based characteristics, to accurately capture and represent structural patterns for a specific downstream task. In specific tasks, such as fraud detection, the required structural patterns can be highly complex. A well-balanced combination of multiple indicators is essential, not only for the focal node but also for the nodes in its $k$-hop neighborhood. The edges of the similarity graph in ffstruc2vec are assigned weights derived from the similarity measures of the node pairs. They are utilized in random walks on the similarity graph to generate sequences of nodes. These sequences are then fed into the word2vec algorithm \citep{Mikolov.2013} to learn embedded node representation vectors. A detailed description of the ffstruc2vec framework can be found in Section~\ref{sec:The_ffstruc2vec_Framework}.

The name ffstruc2vec reflects the framework’s central design principle: a flat similarity graph that flexibly captures structural identities. This architecture enables effective adaptation to diverse downstream tasks by optimizing the weights of graph indicators across multiple $k$-hop neighborhoods. These optimized weights not only guide the embedding process but also offer interpretability by revealing task-relevant structural patterns. Beyond its flexibility and interpretability, ffstruc2vec is designed for scalability, making it well-suited for large-scale networks. Its optimized flat architecture and flexible alignment to task-specific structural requirements allow ffstruc2vec to capture a broader range of structural identities than alternative methods such as struc2vec \citep{Ribeiro.2017}. A direct comparison to struc2vec is presented in Section~\ref{sec:Comparison_of_ffstruc2vec_with_struc2vec}, while a detailed discussion of the general key contributions of ffstruc2vec is provided in Section~\ref{sec:Key_Contributions_and_Advantages_of_ffstruc2vec}.

The effectiveness of ffstruc2vec is demonstrated in multiple downstream application tasks:
\begin{itemize}
	\item Unsupervised applications (Sections \ref{sec:Zachary's_Karate_Club_unsupervised} and \ref{sec:Barbell_graph}) showcase its superior performance.
	\item Supervised tasks (Section~\ref{sec:Air-traffic_network}) highlight its significant improvements over other node embedding methods.
\end{itemize}

The following list provides an overview of our key contributions achieved with ffstruc2vec:
\begin{itemize}
	\item We propose a flexible and flat Node Embedding framework that generates latent representations for nodes. This framework effectively aligns with various types of structural patterns in a graph, making it suitable for a wide range of downstream application tasks (see Section~\ref{sec:Flexibility}).
	\item In terms of explainability and interpretability, the alignment process of ffstruc2vec to a downstream application task provides insight into the graph structures of application tasks and their impact, meaning, and relevance for the examined application scenarios (see Section~\ref{sec:Explainability_and_Interpretability}).
	\item To maintain scalability for large graphs, such as social networks with billions of nodes and edges, the time complexity of ffstruc2vec is 
$\mathcal{O}(\max(|E|, |V| \cdot \log |V|))$ 
for the extraction of certain structural identities, as detailed in Section~\ref{sec:Scalability}.
	\item The ffstruc2vec framework addresses weaknesses in existing node embedding frameworks, such as struc2vec, as detailed in Section~\ref{sec:Comparison_of_ffstruc2vec_with_struc2vec}.
	\item The ffstruc2vec framework significantly outperforms other state-of-the-art node embedding frameworks in several practical downstream application tasks (see Section~\ref{sec:Experimental_evaluation_and_benchmarking}).
	\item In addition to extracting structural patterns, the ffstruc2vec framework can incorporate node proximity properties and node features (see Appendix~\ref{sec:Integration_of_local_structural_properties}).
\end{itemize}

\begin{figure}[H]
\centering
\includegraphics[width=119mm]{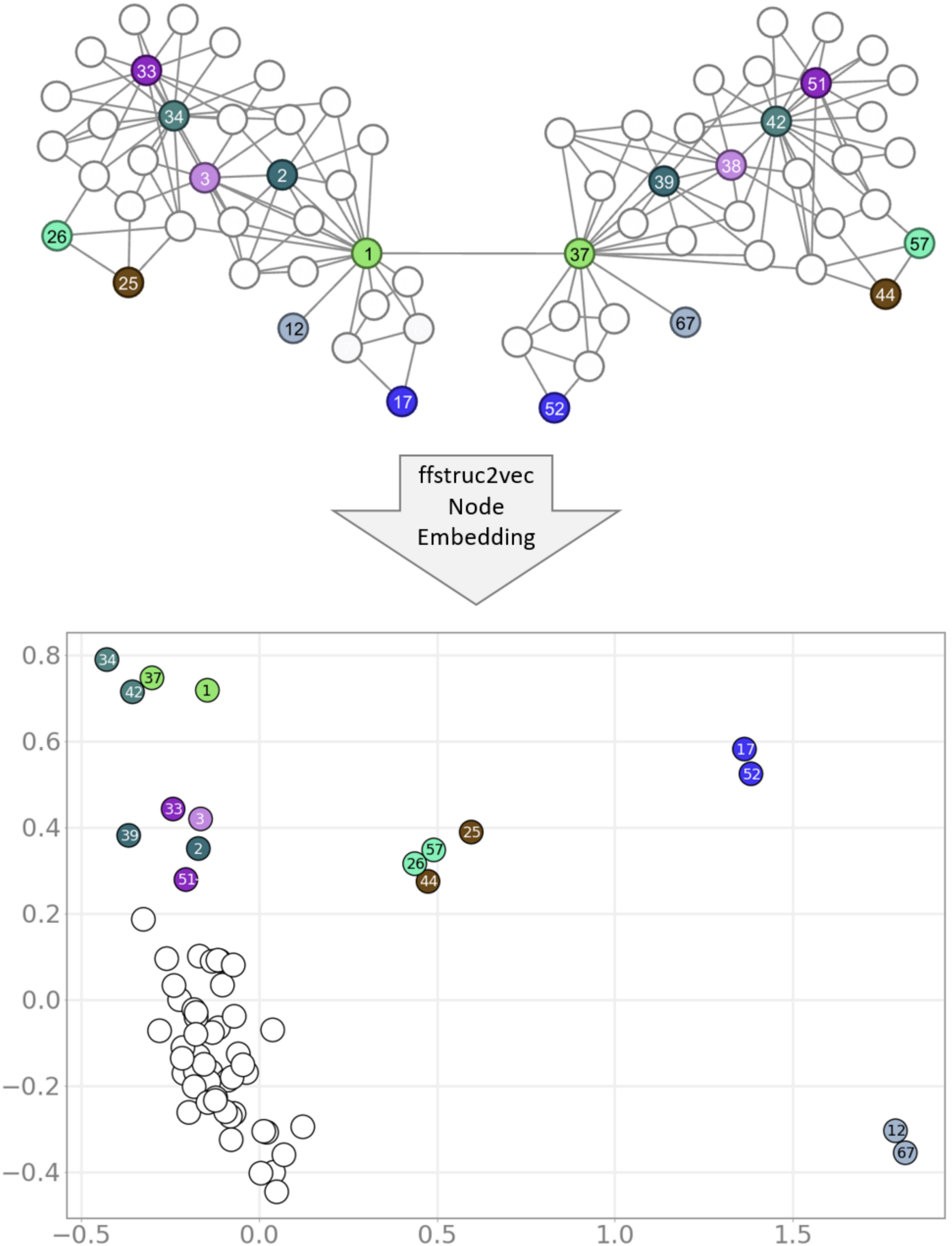}
\caption{ffstruc2vec applied to the mirrored \textit{Zachary's Karate Club}. Nodes with special structural properties in the graph (upper picture), such as central and peripheral nodes, are marked in color. The ffstruc2vec approach effectively separates the embedding vectors of these nodes from the other nodes (lower picture)}\label{fig:figure_zachary_both}
\end{figure}

Figure~\ref{fig:figure_zachary_both} illustrates the application of ffstruc2vec to the mirrored \textit{Zachary's Karate Club} network. The upper image depicts the network, while the lower image visualizes the generated embedded vectors in a two-dimensional space. Nodes with distinct structural properties, such as central and peripheral nodes, are highlighted in color. It is evident that the ffstruc2vec approach effectively separates the embedding vectors of these nodes from the others, demonstrating its ability to distinguish structural roles.

\flushbottom

\section{Related Work}
\label{sec:Related_Work}

Advancements in representational learning for natural language processing opened new ways for feature learning of discrete objects such as words. The Skip-gram model, introduced by \citet{Mikolov.2013}, learns continuous word feature representations by optimizing a neighborhood-preserving likelihood objective. Inspired by this model, \citet{Perozzi.2014} proposed DeepWalk as an analogy for networks by treating a network as a “document”. In the same way that a document is an ordered sequence of words, sequences of nodes can be sampled from a network and transformed into an ordered sequence of nodes. Random walks generate these sequences, and language models then learn node representations by treating the sequences as sentences. Building on this, \citet{Grover.2016} introduced the node2vec approach, which offers a more flexible biased second-order random walk model, while \citet{Ribeiro.2017} proposed struc2vec, focusing on global structural identities. However, struc2vec has several limitations, which are effectively addressed by ffstruc2vec (see Section~\ref{sec:Comparison_of_ffstruc2vec_with_struc2vec}).

Beyond random walks, other node embedding frameworks have been developed to learn low-dimensional representations of nodes in graphs. Most of these frameworks focus on preserving node proximity properties within the resulting embeddings \citep{Perozzi.2014, Grover.2016, Tang.05182015, Hamilton.2017, Qiu.02022018, Lotfalizadeh.15.12.202318.12.2023, Kutzkov.2023, Yin.2023}. However, there has been a growing interest in approaches that focus on structural patterns of nodes, independent of their proximity to others. Consequently, diverse methodologies have emerged.

An example of such an approach is struc2gauss, introduced by \citet{Pei.2020}, which extends structure-based embeddings by representing nodes as Gaussian distributions instead of fixed vectors, capturing uncertainty and hierarchical similarities within structural roles. While struc2gauss models structural uncertainty probabilistically, ffstruc2vec takes a different approach by employing a flat similarity graph structure, which provides high flexibility in adapting to specific downstream application tasks and thereby offers interpretability while maintaining scalability (see Section~\ref{sec:Key_Contributions_and_Advantages_of_ffstruc2vec}).

Additional methods for generating node representations, often integrated into broader embedding frameworks, include self-attention-based approaches \citep{Nguyen.2021}, attribute-aware sampling methods \citep{Li.2024}, kernel-based methods \citep{Nikolentzos.2019}, recursive methods \citep{Tu.07192018}, matrix multiplication methods \citep{Henderson.2012, DiJin.2019, Ouyang.2023}, spectral graph wavelet approaches \citep{Donnat.2018}, contrastive learning methods, including Graph Contrastive Learning approaches \citep{Velickovic.2019, Zhang.2024, Shao.2024, Yang.2024}, and methods targeting specific structural properties, such as role-based embeddings \citep{Ahmed.2022}, anonymous walks \citep{Ivanov.2018b, Qiu.2022, Yan.10212024}, graphlets \citep{Lyu.11062017, Dutta.2019, Tu.2020}, or self-defined ripple diffusion patterns \citep{Luo.2022}. For a broader and more in-depth review of node embedding techniques and their applications, we refer the reader to existing surveys \citep{Cai.2018, Jin.2022, Khoshraftar.2024}.

Each of these methods has its limitations. For example, matrix multiplication-based frameworks tend to be computationally intensive, struggle with flexibility regarding diverse structural patterns, and rely on assumptions about the relationship between the underlying network structure and the downstream application task. The diffusion wavelet approach in GraphWave produces embeddings that are less intuitive in terms of direct structural features and can be costly to compute at scale. Some methods explicitly assign role distributions or structural types to nodes \citep{Henderson.2012, Ahmed.2022}, which can lead to reduced granularity and limit the ability to reflect nuanced structural differences in the resulting embeddings. Deep learning-based approaches, particularly those relying on Graph Neural Networks (GNNs), often face challenges in interpretability and scalability. While some GNN variants attempt to incorporate structural features explicitly—such as Gralsp \citep{Jin.2020}—they require extensive labeled data and typically remain tightly coupled to specific downstream tasks, which also limits their reusability. In contrast, Graph Contrastive Learning aims to produce task-agnostic representations in a self-supervised fashion, but relies on computationally demanding data augmentation or multi-view training pipelines, which further limit its scalability and adaptability in application-specific settings.

In contrast, \texttt{ffstruc2vec} addresses these limitations by providing task-adaptive and transparent embeddings within a design that remains efficient even on large-scale graphs. To the best of our knowledge, no existing framework offers a comparable combination of downstream-task flexibility, interpretability, and scalability (see Section~\ref{sec:Key_Contributions_and_Advantages_of_ffstruc2vec}).

\section{Definitions and Preliminaries}
\label{sec:Definitions_and_Preliminaries}
This section defines the fundamental definitions and preliminary concepts required in this paper.

\begin{definition}\label{def:Graph} 
(Graph) A \textit{graph}, denoted as $G = (V,E)$, is a mathematical construct comprising a set of vertices $V = \{ v_1, \dots, v_n \}$ and a set of edges $E \subseteq \big\{ \{ u, v \} \mid u, v \in V, u \neq v \big\}$ in the case of an \textit{undirected graph}, meaning each edge is an unordered pair of distinct nodes. In contrast, in a \textit{directed graph}, the edge set is defined as $E \subseteq \big\{ (u, v) \mid u, v \in V, u \neq v \big\}$, where edges are ordered pairs indicating a directed connection from node $u$ to node $v$.

In this work, we exclusively focus on undirected graphs, as defined above. However, the fundamental concepts introduced in this paper can be easily adapted and extended to directed graphs.
\end{definition}

\begin{definition}\label{def:Node_embedding} 
(Node Embedding) A \textit{Node Embedding} of a graph $G=(V,E)$ with $n$ nodes is defined as a mapping function 
\[
f: v_i \mapsto y_i \in \mathbb{R}^d, \quad \forall i \in \{1, \dots, n\},
\]
where $d \ll |V|$, and the function $f$ preserves certain properties defined on the graph $G$.
\end{definition}

\begin{figure}[h]
\centering
\includegraphics[width=119mm]{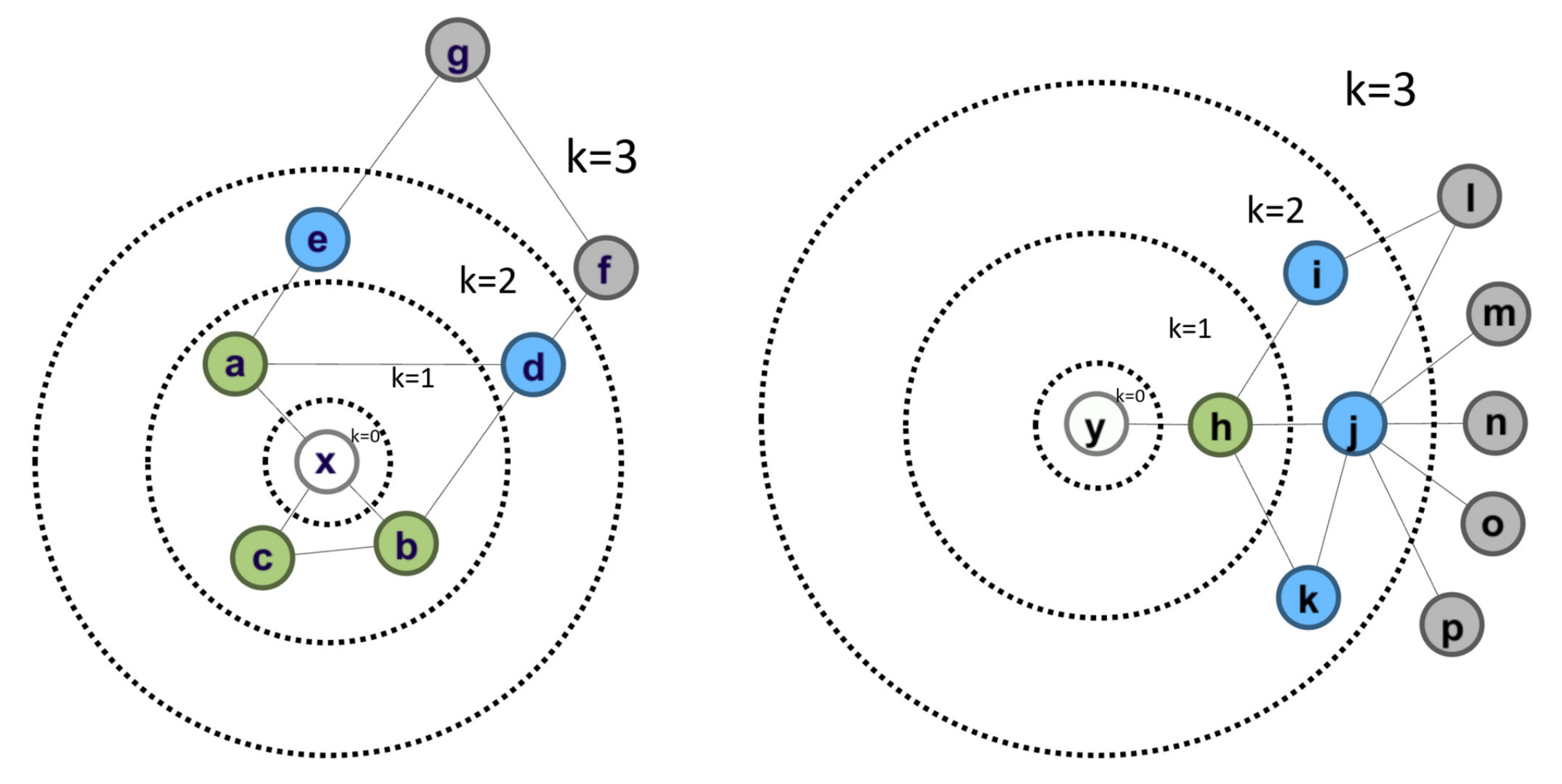}
\caption{Visualization of $k$-hop neighborhoods for nodes $x$ and $y$. The neighborhoods are color-coded as follows: green for $k = 1$, blue for $k = 2$, and gray for $k = 3$}\label{fig:fig2}
\end{figure}

\begin{definition}\label{def:k-hop_neighborhoods} 
($k$-hop Neighborhood) The \textit{$k$-hop neighborhood} of a node $x$ in a graph is the set of all nodes that are exactly $k$ hops away from $x$. Formally, it is defined as
\begin{equation*}
    N_k(x) = \{ y \in V \mid \text{distance}(x,y) = k \},
\end{equation*}
where $V$ represents the set of all nodes in the graph, and $\text{distance}(x,y)$ represents the length of the shortest path between nodes $x$ and $y$. 

For an intuitive understanding, Figure~\ref{fig:fig2} visually illustrates the \textit{$k$-hop neighborhoods} of two nodes, $x$ and $y$, using color coding. Specifically, the \textit{1-hop neighborhood} is highlighted in green, the \textit{2-hop neighborhood} in blue, and the \textit{3-hop neighborhood} in gray. As an illustrative example, the \textit{2-hop neighborhoods} of nodes $x$ and $y$ are given by
\begin{equation*}
    N_2(x) = \{ e,d \}, \quad N_2(y) = \{ i,j,k \}.
\end{equation*}
\end{definition}

\begin{definition}\label{def:node_degree} 
(Node Degree) In accordance with Definition~\ref{def:Graph}, the \textit{degree of a node} $v$ in a graph $G=(V,E)$ is defined as
\begin{equation*}
    d_v = \big| \{ u \mid \{ v,u \} \in E \} \big|.
\end{equation*}
\end{definition}

\begin{definition}\label{def:isomorphic_Graphs} 
(Isomorphic Graphs) Two graphs $G = (V,E)$ and $G' = (V',E')$ are \textit{isomorphic} if there exists a bijective mapping $g: V \to V'$ (called the isomorphism function) such that for any two nodes $v_i, v_j \in V$, the following condition holds:
\begin{equation*}
    (v_i,v_j) \in E \quad \text{iff} \quad (g(v_i), g(v_j)) \in E'.
\end{equation*}
\end{definition}

\begin{figure}[h]
\centering
\includegraphics[width=119mm]{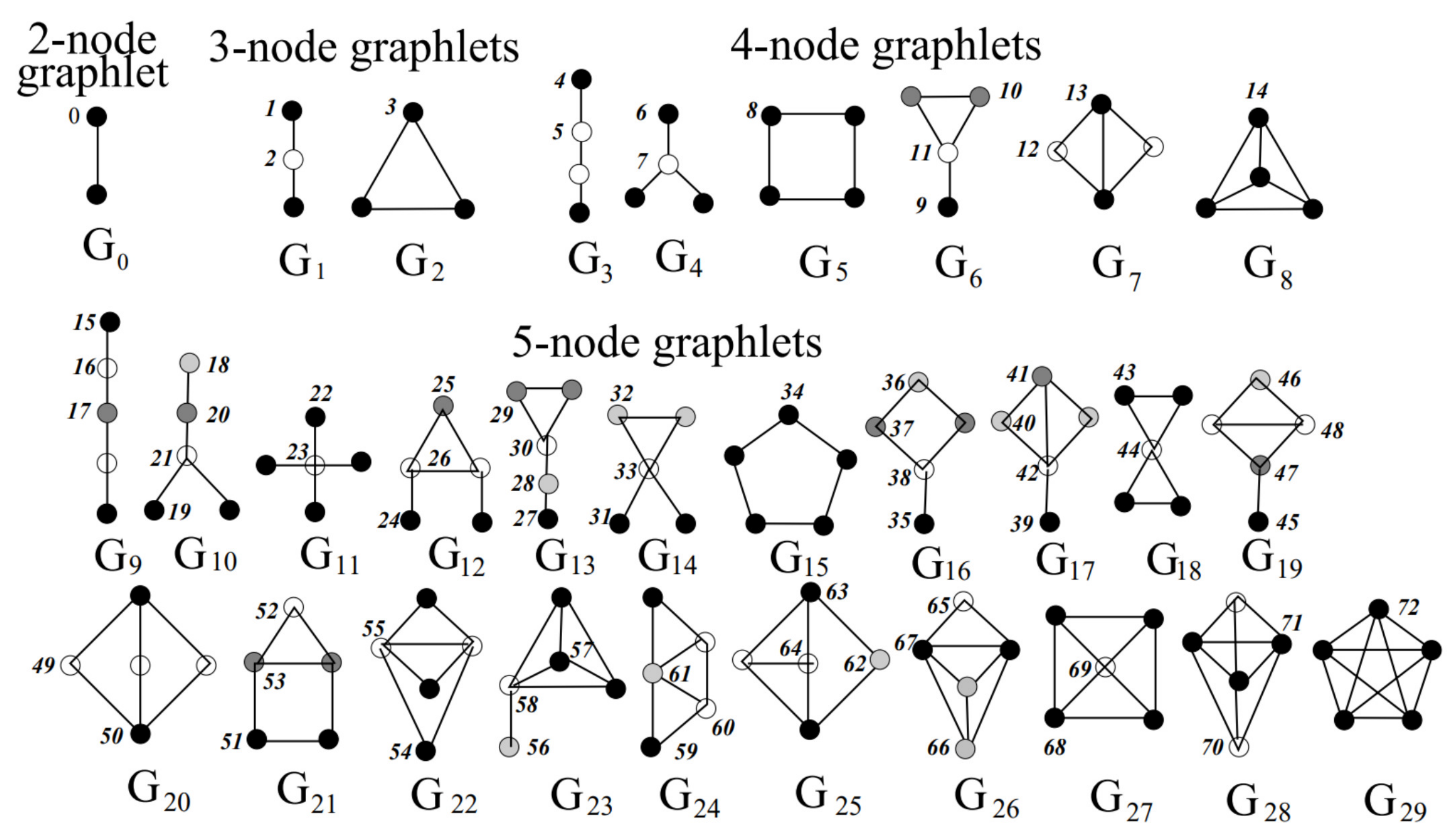}
\caption{Automorphism orbits $0, 1, 2, \dots, 72$ for the thirty $2$-, $3$-, $4$-, and $5$-node graphlets $G_0, G_1, \dots, G_{29}$. In a graphlet $G_i$ for $i \in \{0,1,\dots,29\}$, nodes belonging to the same orbit are shaded identically \citep{Przulj.2007}}
\label{fig:fig3}

\end{figure}

\begin{definition}\label{def:graphlets_and_orbits} (Graphlets and Orbits) \textit{Graphlets} are defined as small, connected, and non-isomorphic induced subgraphs (as specified in Definition~\ref{def:isomorphic_Graphs}) of a graph, consisting of at least two nodes \citep{Przulj.2004}. All \textit{graphlets} with $2$, $3$, $4$, or $5$ nodes and their automorphism \textit{orbits} are illustrated in Figure~\ref{fig:fig3}. \textit{"orbit"} refers to the different possible node positions within the \textit{graphlets}. Symmetrical nodes are assigned the same orbit number, with $73$ orbits of $30$ \textit{graphlets}.
\end{definition}

\begin{definition}\label{def:Anonymous_Walk} 
(Anonymous Walk) Let $s = (u_1, u_2, \dots, u_k)$ be an ordered list of elements $u_i \in V$. We define the positional function $pos: (s, u_i) \mapsto q$, which returns a list $q = (p_1, p_2, \dots, p_l)$ of all positions $p_j \in \mathbb{N}$ where $u_i$ occurs in the list $s$. 

If $w = (v_1, v_2, \dots, v_k)$ is a random walk, then its corresponding \textit{anonymous walk} is the sequence of integers $a = (f(v_1), f(v_2), \dots, f(v_k))$, where the function $f$ is defined as
\begin{equation*}
    f(v_i) = \min_{p_j \in pos(w, v_i)} pos(w, v_i).
\end{equation*}
\end{definition}

\section{The ffstruc2vec Framework}
\label{sec:The_ffstruc2vec_Framework}

The ffstruc2vec framework learns structural node representations by capturing similarities between nodes based on their structural roles within a network. It computes structural similarity scores between node pairs using a diverse set of structural properties, both of individual nodes and their $k$-hop neighborhoods. The framework provides a flexible mechanism for integrating various comparison functions and adaptive weightings (see Section~\ref{sec:Measure_structural_similarity}), allowing for an effective and interpretable encoding of structural identities. Based on these similarities, a similarity graph is constructed, where edge weights quantify the computed relationships (see Section~\ref{sec:Construct_the_similarity-graph}). To extract meaningful patterns from this graph, biased random walks are performed, with transition probabilities influenced by the similarity-based edge weights (see Section~\ref{sec:Performing_biased_random_walks}). The resulting node sequences are then used in the Skip-gram model to learn dense vector representations of nodes (see Section~\ref{sec:Generate_node_representation_vectors}). Finally, a task-aware optimization step refines the embeddings to improve performance in specific downstream applications (see Section~\ref{sec:Task_Aware_Optimization_of_Node_Representations}).

A formalized overview of the complete ffstruc2vec pipeline is provided in Algorithm~\ref{algo:ffstruc2vec}, summarizing all steps in a structured manner.

\begin{algorithm}
\caption{ffstruc2vec: Learning Pipeline}
\label{algo:ffstruc2vec}
\textbf{Input:} Graph $G(V, E)$, maximum neighborhood depth $k^*$ (defining the maximum $k$-hop neighborhood),  
$l$ graph indicators $\{I_i\}_{i=1}^{l}$ (see Appendix~\ref{sec:Graph_indicators}),  
$l$ comparison functions $\{f_i\}_{i=1}^{l}$ (see Appendix~\ref{sec:Comparison_of_the_graph_indicators_of_node_groups}),  
$p$: number of random walks per node,  
$L_w$: length of each random walk,  
$d$: embedding dimension.  

\textbf{Output:} Node representation vectors $\{r_x\}_{x \in V}$ in $\mathbb{R}^d$.  

\begin{algorithmic}[1]
    \ForAll{$x \in V$}
        \State Compute graph indicators $\{I_i(x)\}_{i=1}^{l}$ \hfill // see Appendix~\ref{sec:Graph_indicators}
    \EndFor
   
   \State Initialize weights $w_{ik}$ for all $(i, k) \in \{1, \dots, l\} \times \{0, \dots, k^*\}$
   
    \Statex \hspace{2.5em} for all $(i, k) \in \{1, \dots, l\} \times \{0, \dots, k^*\}$

    \ForAll{$(x, y) \in V \times V$}
        \State $sim(x, y) \gets$ computeStructuralSimilarity
        \Statex \hspace{2.5em} $\big(x, y, \{I_i(x)\}_{i=1}^{l}, \{f_i\}_{i=1}^{l}, \{w_{ik}\} \big)$
        \hfill // see Section~\ref{sec:Measure_structural_similarity}
    \EndFor

    \State $G'(V', E') \gets$ similarityGraph$(V, sim(x, y))$ \hfill // see Section~\ref{sec:Construct_the_similarity-graph}

    \ForAll{$x \in V$}
        \State $context(x) \gets$ performRandomWalks$(G', x, p, L_w)$ \hfill // see Section~\ref{sec:Performing_biased_random_walks}
    \EndFor

    \State $r \gets$ word2vec$\left(\bigcup_{x \in V} \text{context}(x), d\right)$ \hfill // see Section~\ref{sec:Generate_node_representation_vectors}

\State Optimization of the task-specific objective by aligning
\Statex \hspace{1.5em} ffstruc2vec to the structural-specific requirements
\Statex \hspace{1.5em} of the respective application task, primarily through
\Statex \hspace{1.5em} task-aware weight adjustments, if applicable.
\hfill // see Section~\ref{sec:Task_Aware_Optimization_of_Node_Representations}

\end{algorithmic}
\end{algorithm}

\subsection{Measuring Structural Similarity}
\label{sec:Measure_structural_similarity}

The ffstruc2vec framework begins by evaluating the structural similarity between pairs of nodes in a graph, leveraging graph indicators as metrics to quantify the structural properties of individual nodes. Notably, the similarity between a node $x$ and a node $y$ is not solely determined by their respective structural properties but also accounts for the structural characteristics of the nodes within their $k$-hop neighborhoods, $N_k(x)$ and $N_k(y)$ (see Definition~\ref{def:k-hop_neighborhoods}).

As an initial step, the structural similarity of the $k$-hop neighborhoods of $x$ and $y$, denoted as $sim_k(x,y)$, is computed by comparing the graph indicator values of the nodes in $N_k(x)$ with those of the nodes in $N_k(y)$. Let $I = \{ I_1, I_2, \dots, I_l \}$ be a set of graph indicators $I_i: V \to W(I_i)$, where each indicator describes a specific structural property of a node, and \( W(I_i) \) denotes the value range of the indicator. Examples of such graph indicators are provided in Appendix~\ref{sec:Graph_indicators}.

To quantify the similarity, let 
\begin{equation*}
    f_i: \mathcal{P}(W(I_i)) \times \mathcal{P}(W(I_i)) \to \mathbb{R}, \quad i \in \{ 1, \dots, l \},
\end{equation*}
denote a set of functions that compare the values of the graph indicators between two sets of nodes, where \( \mathcal{P}(W(I_i)) \) denotes the power set of \( W(I_i) \), i.e., the set of all subsets of the value range of the graph indicator \( I_i \). Specific examples of these comparison functions are detailed in Appendix~\ref{sec:Comparison_of_the_graph_indicators_of_node_groups}. The structural similarity $sim_k(x,y)$ is then computed as defined in Equation~\ref{eq:simk_xy}.

\begin{equation}\label{eq:simk_xy}
sim_k(x,y) = \sum _{i=1}^{l} w_{ik} \cdot f_i \left( \bigcup\limits_{v\in N_k(x)} \{ I_i(v) \}, \bigcup\limits_{v\in N_k(y)} \{ I_i(v) \} \right)
\end{equation}
where $w_{ik}\in \mathbb{R}$ are weighting factors applied to the values of graph indicators $I_i$ for the $k$-hop neighborhoods.

In the subsequent step of the ffstruc2vec framework, the overall structural similarity $sim(x,y)$ between nodes $x$ and $y$ is computed as the sum of the structural similarities between their respective $k$-hop neighborhoods, denoted as $sim_k(x,y)$, for all $k\leq k^*$. Here, $k^*$ is a predefined constant that determines the maximum neighborhood depth considered when describing the structural properties of a node’s neighborhood (see Equation~\ref{eq:sim_xy}).

\begin{equation}\label{eq:sim_xy}
sim(x,y)=\sum _{k=0}^{k^*} sim_k(x,y)
\end{equation}

\noindent
By substituting Equation~\ref{eq:simk_xy} into Equation~\ref{eq:sim_xy}, we obtain the following expanded expression.

\begin{equation}\label{eq:sim_xy_long}
sim(x,y) = \sum _{k=0}^{k^*} \sum _{i=1}^{l} w_{ik} \cdot f_i\left(\bigcup\limits_{v\in N_k(x)} \{ I_i(v) \}, \bigcup\limits_{v\in N_k(y)} \{ I_i(v) \} \right)
\end{equation}

\noindent
For notational convenience, we extend the function $I_i$ to sets by defining

\begin{equation}\label{eq:extended}
I_i(N_k(x)) := \bigcup\limits_{v\in N_k(x)} \{ I_i(v) \}
\end{equation}

\noindent
Using Equation~\ref{eq:extended}, we can rewrite Equation~\ref{eq:sim_xy_long} in a more compact form as follows.

\begin{equation}\label{eq:sim_xy_final}
sim(x,y) = \sum _{k=0}^{k^*} \sum _{i=1}^{l} w_{ik} \cdot f_i(I_i(N_k(x)), I_i(N_k(y)))
\end{equation}

\noindent
The weighting of graph indicators in different $k$-hop neighborhoods using the weighting factors $w_{ik}$ in Equation~\ref{eq:sim_xy_final} is a key aspect of ffstruc2vec's flexibility (see Section~\ref{sec:Task_Aware_Optimization_of_Node_Representations} and Section~\ref{sec:Flexibility}) in extracting diverse types of structural identities tailored to various downstream application tasks.

\begin{remark}\label{remark:optimization}
When employing a large number of graph indicators $(l)$ and multiple $k$-hop neighborhoods $(k^*+1)$, the search for $l\cdot (k^*+1)$ weighting factors suitable for a specific downstream application task—e.g., through heuristic optimization—can be computationally intensive. The following strategies help mitigate the computational cost of determining suitable weighting factors.

\begin{itemize}
	\small
	\item In practical applications, a choice of $k^*\leq 4$ is reasonable for most downstream application tasks. 
	This is supported by empirical research on the Facebook network of active users, which found an average shortest path length of $4.74$ between two nodes \citep{Backstrom.06222012}.
	\item To reduce the number of flexible weighting factors from $l \cdot (k^* + 1)$ to $l + k^* + 1$, the weighting factors for graph indicators and $k$-hop neighborhoods can be integrated independently, as shown in Equation~\ref{eq:sim_xy_reduce}. While this modification improves the computational efficiency of ffstruc2vec, it comes at the expense of a slight reduction in expressive power.

\begin{equation}\label{eq:sim_xy_reduce}
sim(x,y) = \sum _{k=0}^{k^*} w_k^{\text{hop}} \sum _{i=1}^{l} w_i^{\text{ind}} \cdot f_i(I_i(N_k(x)), I_i(N_k(y)))
\end{equation}
	\item To determine $l\cdot (k^*+1)$ suitable weighting factors for large values of $l$ and $k^*$, a smaller graph subset can be used during the optimization phase before applying ffstruc2vec to the full graph with the optimized weighting factors. This approach enhances computational efficiency during optimization but increases the risk of selecting weighting factors that may be less optimal for specific downstream application tasks.
\end{itemize}
\end{remark}

\subsection{Constructing the Similarity Graph}
\label{sec:Construct_the_similarity-graph}

As defined in Definition~\ref{def:Graph}, the original graph is denoted as $G=(V,E)$. A fully connected weighted graph $G'=(V',E')$ is constructed, to encode the structural similarities between all nodes in $G$. Specifically, the node set remains unchanged, i.e., $V'=V$, while the edge weights in $E'$ are assigned for $(x,y)\in E'$ based on the structural similarity between nodes $x$ and $y$ in the original graph $G$, quantified by $sim(x,y)$ in Equation~\ref{eq:sim_xy_final}.

The edge weights are computed by applying a transformation function to $sim(x,y)$, ensuring that structurally similar nodes receive higher weights. The choice of this function in $G'$ is designed to be flexible and is treated as a hyperparameter (see Appendix~\ref{sec:Hyperparameters}). This adaptability allows the weighting function to be customized for the specific requirements of a given downstream application task.

\begin{figure}[h]
\centering
\includegraphics[width=119mm]{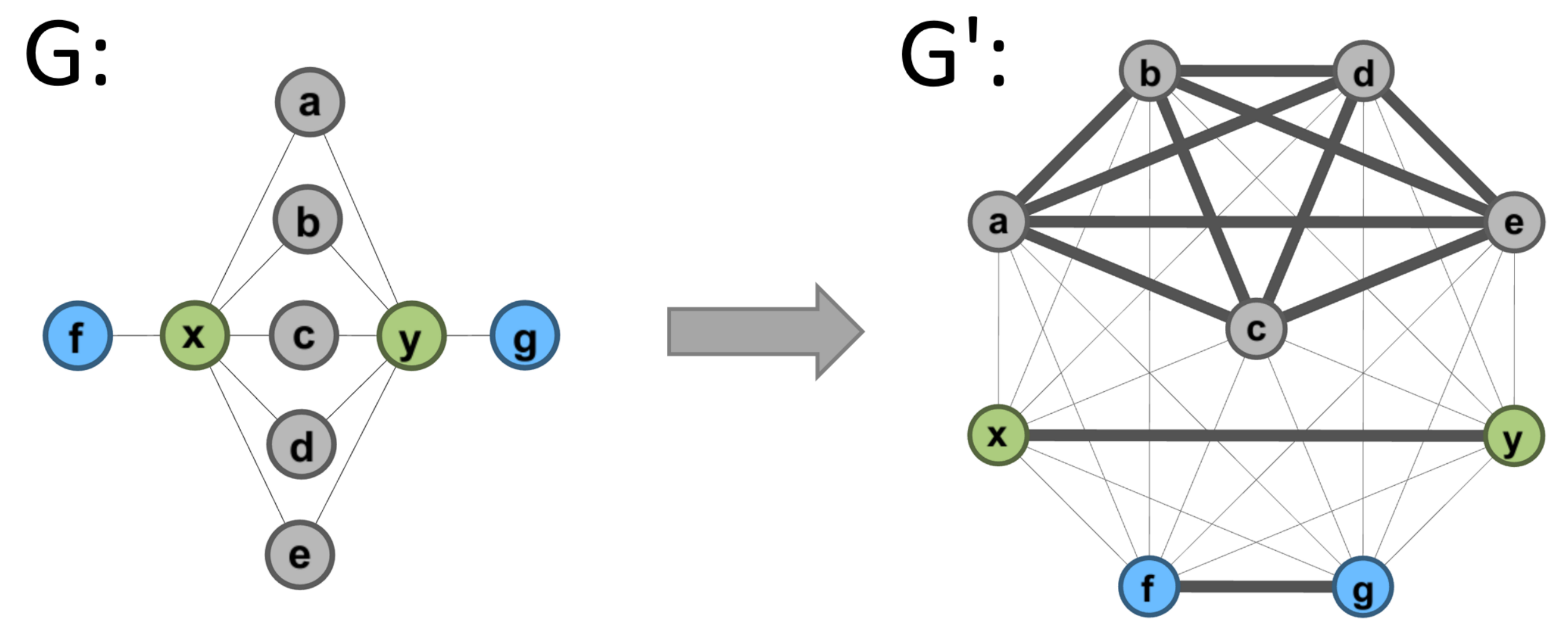}
\caption{Example of a similarity graph $G'$ constructed from the original graph $G$. Nodes with the same structural properties in $G$ are depicted in the same color, while edges with high weights—indicating a high degree of structural similarity between the connected nodes—are represented as thick lines in $G'$}
\label{fig:fig4}
\end{figure}

An illustrative example of this process is provided in Figure~\ref{fig:fig4}. Nodes in $G$ with identical structural properties are depicted in the same color. In $G'$, edges connecting these structurally similar nodes are assigned higher weights, which are visually represented as thicker lines.

\subsection{Performing biased random walks}
\label{sec:Performing_biased_random_walks}

To generate node representation vectors using the similarity graph $G'$, we first perform a biased random walk on $ G'$. During this process, the selection of the next node to visit is determined by the edge weights in $G'$, ensuring that transitions favor structurally similar nodes.  

The transition probability from node $x$ to node $y$, denoted as $P_r(x \to y)$, is computed as:
\begin{equation}\label{eq:prob}
P_r(x \to y) = \frac{w_{xy}}{\sum_{j=1}^{n} w_{xj}}
\end{equation}
where $n$ represents the total number of nodes in the graph, and $w_{xy}$ denotes the edge weight of the edge connecting nodes $x$ and $y$ in $G'$.

\begin{figure}[h]
\centering
\includegraphics[width=119mm]{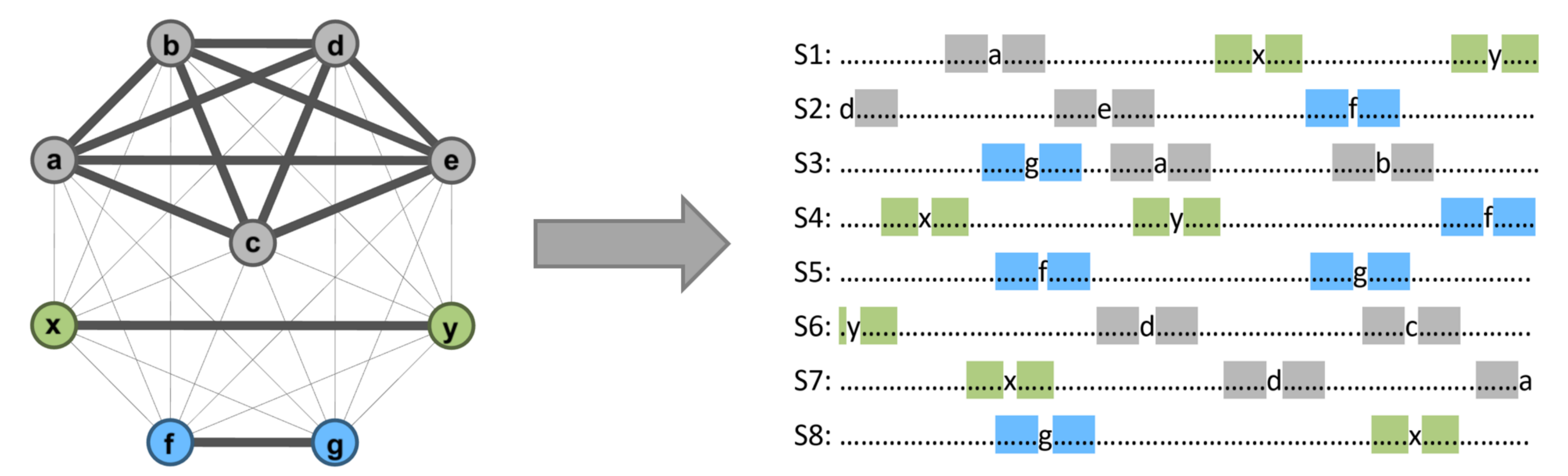}
\caption{Generating node sequences (right) using random walks on the similarity graph (left). Structurally similar nodes in the original graph result in similar contexts within the generated sequences, as indicated by the matching colors in the illustrations}
\label{fig:fig5}
\end{figure}

Multiple random walks of a specified length are initiated from each node, generating sequences of nodes. The selection of the next node in these walks follows the transition probabilities defined in Equation~\ref{eq:prob}, ensuring that transitions occur based on the edge weights in $G'$. Since the edge weights in the fully connected graph $G'$ encode the structural similarities between nodes in $G$ (as described in Section~\ref{sec:Construct_the_similarity-graph}), the random walks naturally tend to position structurally similar nodes in similar contexts within the generated node sequences (as illustrated in Figure~\ref{fig:fig5}).

\subsection{Generating Node Representation Vectors}
\label{sec:Generate_node_representation_vectors}

\begin{figure}[h]
\centering
\includegraphics[width=119mm]{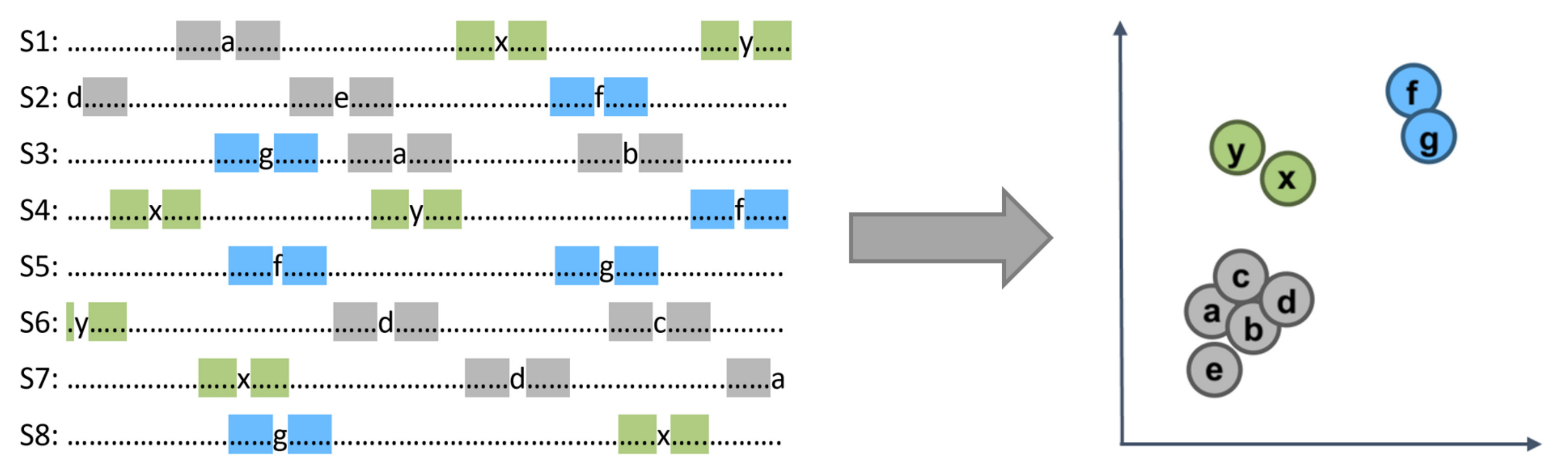}
\caption{Applying the Skip-gram model to the generated node sequences (left) to learn node representation vectors (right). Nodes with similar structural properties are assigned similar colors, reflecting their embedding in similar contexts within the sequences}
\label{fig:fig6}
\end{figure}

The Skip-gram language model \citep{Mikolov.2013} is employed to generate similar representation vectors for nodes that appear in similar contexts within these sequences. Since similar structural identities in $G$ produce similar node contexts in the sequences, the model learns low-dimensional node representation vectors in a continuous vector space, ensuring that structurally similar nodes in $G$ yield similar representations (as illustrated in Figure~\ref{fig:fig6}).  

More specifically, the Skip-gram model learns a function $\Phi$ that maps nodes to their representation vectors by maximizing the conditional probability
\[
P_r\left(\{ v_{i-w}, \dots, v_{i-1}, v_{i+1}, \dots, v_{i+w} \} \mid \Phi(v_i) \right),
\]
where the context of node $v_i$ in a sequence is defined as a window of size $w$ centered on the node.

\subsection{Task-Aware Optimization of Node Representations}
\label{sec:Task_Aware_Optimization_of_Node_Representations}

Different downstream application tasks rely on different types of structural identities within a network, as discussed in Section~\ref{sec:Related_Work} and Section~\ref{sec:Flexibility}. ffstruc2vec is highly flexible, allowing for the seamless integration of various graph indicators with corresponding selectable comparison functions (see Appendices~\ref{sec:Graph_indicators}--\ref{sec:Comparison_of_the_graph_indicators_of_node_groups}). Moreover, it enables the task-specific weighting of these structural properties for each graph indicator $i$, both for the nodes themselves and for each of their $k$-hop neighborhoods, using the parameters $w_{ik}$ for all $(i, k) \in \{1, \dots, l\} \times \{0, \dots, k^*\}$ (see Section~\ref{sec:Measure_structural_similarity}), where $k^*$ defines the maximum neighborhood depth under consideration.

These parameters can either be set manually or, if a downstream application task is accompanied by a loss function, optimized accordingly to align with the task-specific objective. A straightforward approach is to apply heuristic methods such as evolutionary strategies or Bayesian optimization techniques—for example, the Tree-structured Parzen Estimator (TPE) \citep{Bergstra.2011}. For large datasets with high values of $l$ and $k^*$, the measures discussed in Remark~\ref{remark:optimization} can be applied to ensure scalability.

\section{Key Contributions and Advantages of ffstruc2vec}
\label{sec:Key_Contributions_and_Advantages_of_ffstruc2vec}

\subsection{Flexibility in Extracting Diverse Structural Identities}
\label{sec:Flexibility}

The structural identity of a node can be described through various types of structural patterns. The type of structural patterns relevant to a downstream application task may vary depending on the specific case. To ensure the effective application of node embedding vectors, they must preserve the types of structural patterns relevant to the given task. Therefore, a node embedding framework has to possess the capability to adapt flexibly to the specific requirements of application tasks. Many existing approaches lack this flexibility (see Section~\ref{sec:Related_Work}).

ffstruc2vec provides this flexibility through several hyperparameters outlined in more detail from Appendix~\ref{sec:Graph_indicators} to Appendix~\ref{sec:Hyperparameters}. It can combine several graph indicators describing structural properties (see Appendix~\ref{sec:Graph_indicators}), node proximity measures, and node features (see Appendix~\ref{sec:Integration_of_local_structural_properties}). Moreover, it allows adjusting the structural focus not only on the node itself but also on different layers of its $k$-hop neighborhood. The various layers of the $k$-hop neighborhoods of nodes, as defined in Definition~\ref{def:k-hop_neighborhoods}, have varying levels of importance for different application tasks. While some tasks require extracting specific structural properties of the nodes themselves, others may require the identification of specific structural properties of the immediate or the indirect neighborhood to make accurate inferences. Therefore, ffstruc2vec assigns weights to each graph indicator in each $k$-hop neighborhood to meet the requirements of specific application tasks, as outlined in Section~\ref{sec:Measure_structural_similarity}. To compare the structural identities of two nodes, various comparison functions can be applied (see Appendix~\ref{sec:Comparison_of_the_graph_indicators_of_node_groups}) that meet the requirements of the task at hand and enable the seamless incorporation of even complex structural properties. Furthermore, ffstruc2vec provides additional flexibility through a spectrum of other hyperparameters, as detailed in Appendix~\ref{sec:Hyperparameters}.

Optimized hyperparameters can be determined through various methods. Depending on the requirements of the application task, various evaluation metrics can be employed to assess the effectiveness of the selected hyperparameters and the associated structural patterns. The continuous nature of the weights results in a vast hyperparameter space. Therefore, we optimized the accuracy score for the classification task described in Section~\ref{sec:Air-traffic_network} using the Tree-structured Parzen Estimator (TPE), as introduced by \citet{Bergstra.2011}. 

The resulting hyperparameter configurations provide valuable insights into the explanation and interpretation of the structural patterns relevant to the application task, as discussed in Section~\ref{sec:Explainability_and_Interpretability}. ffstruc2vec enables this level of explainability and interpretability enabled by its adaptable framework, which formulates the optimization problem in a linear search space (see Section~\ref{sec:Measure_structural_similarity}).

\subsection{Explainability and Interpretability}
\label{sec:Explainability_and_Interpretability}

Explainability and interpretability of machine learning algorithms \citep{Atzmueller.2024, Schwalbe.2024} are becoming increasingly important in many practical applications, especially in highly regulated domains.

By generating structural signatures, ffstruc2vec can provide valuable insights into downstream application tasks. For example, the optimized hyperparameters of the flexible ffstruc2vec framework—such as the weights assigned to graph indicators and $k$-hop neighborhoods (see Section~\ref{sec:Flexibility})—can reveal typical structural patterns associated with the embedding of specific nodes in the context of the application.

An illustrative example is a classification task for detecting financial fraud in transaction graphs, where edges represent financial interactions between entities. Optimized weights may highlight structural motifs, such as triangular patterns or graphlets within a node's neighborhood, as indicative of fraudulent behavior. In the case of an air traffic network (see Section~\ref{sec:Air-traffic_network}), high centrality or bridge scores for an airport or its neighbors may correspond to high passenger volumes. In contrast, in social networks, the same indicators may suggest different behavioral or relational dynamics, depending on the specific application context \citep{Huang.2014}.

To further improve the interpretability of learned weights, preprocessing steps such as normalization or standardization of graph indicators can be beneficial.

\subsection{Scalability}
\label{sec:Scalability}

ffstruc2vec is designed to scale efficiently with the number of nodes and edges in the input graph, making it suitable for real-world scenarios such as large-scale social networks with billions of elements. Optimization measures applied to the algorithm result in the following time complexity for the extraction of certain structural identities:
\[
\mathcal{O}(\max(|E|,\ |V|\cdot\log|V|))
\]
\noindent
Details on the complexity analysis for the optimized version of ffstruc2vec are provided in Appendix~\ref{sec:Complexity_analysis}.

\section{Comparative Analysis of ffstruc2vec and struc2vec}
\label{sec:Comparison_of_ffstruc2vec_with_struc2vec}

In node embedding, struc2vec is a state-of-the-art framework that employs random walks to preserve nodes' structural identities. The following points summarize the advantages of ffstruc2vec over struc2vec, with further details provided in Appendix~\ref{sec:Advantages_of_ffstruc2vec_over_struc2vec}.

\begin{enumerate}

\item[(a)] \textbf{Greater Flexibility}\\ffstruc2vec offers a high degree of flexibility in extracting various types of structural identities, enabling better alignment with specific requirements of downstream application tasks.

\item[(b)] \textbf{Explainability and Interpretability}\\The alignment process of ffstruc2vec with downstream application tasks provides valuable insights into graph structures, highlighting their impact, meaning, and relevance.

\item[(c)] \textbf{Superior Scalability}\\ffstruc2vec is more scalable for large graphs, with a time complexity of $\mathcal{O}(\max(|E|,|V|\cdot\log|V|))$ for the extraction of certain structural identities, compared to struc2vec’s $\mathcal{O}(|V|^3)$ complexity.
	
\item[(d)] \textbf{Optimized Flat Similarity Graph}\\Unlike struc2vec’s multilayer graph, ffstruc2vec employs a flat structure to generate node embeddings, improving efficiency and performance.
	
\item[(e)] \textbf{Fewer Restrictions on Structural Identity Extraction}
struc2vec's design imposes limitations on extracting certain structural identities for specific tasks, whereas ffstruc2vec provides a more adaptable framework.
	
\item[(f)] \textbf{Better Downstream Performance}\\ffstruc2vec significantly outperforms struc2vec on supervised and unsupervised downstream application tasks, as demonstrated in Section~\ref{sec:Experimental_evaluation_and_benchmarking}.

\end{enumerate}

\section{Experimental Evaluation and Benchmarking}
\label{sec:Experimental_evaluation_and_benchmarking}

In this section, ffstruc2vec demonstrates its ability to extract structural identities by applying two unsupervised tasks, detailed in Sections \ref{sec:Zachary's_Karate_Club_unsupervised} and \ref{sec:Barbell_graph}, and a supervised task in Section~\ref{sec:Air-traffic_network}. Specifically, visualization, clustering, and outlier detection are used for unsupervised tasks, while node classification is applied to supervised tasks. Additionally, the performance of ffstruc2vec is evaluated by benchmarking it against other node embedding frameworks.

\subsection{Zachary's Karate Club}
\label{sec:Zachary's_Karate_Club_unsupervised}

The \textit{Zachary's Karate Club} network, originally introduced by \citet{Zachary.1977}, consists of 34 nodes and 78 edges. Each node represents a club member, while the edges denote interactions between members outside the club. These edges are commonly interpreted as indicators of friendship among members.

In this section, we utilize a network previously used by \citet{Ribeiro.2017} to compare ffstruc2vec with the node embedding methods struc2vec, node2vec, and DeepWalk. This network comprises two identical copies of \textit{Zachary's Karate Club} network, designated as $G_1$ and $G_2$. Each node in $G_1$, denoted as $v$, has a corresponding mirror node in $G_2$, denoted as $u$, where $u$ and $v$ share identical structural properties. These properties are expected to be preserved in the node embedding vectors.

\begin{figure}[h]
\centering
\includegraphics[width=119mm]{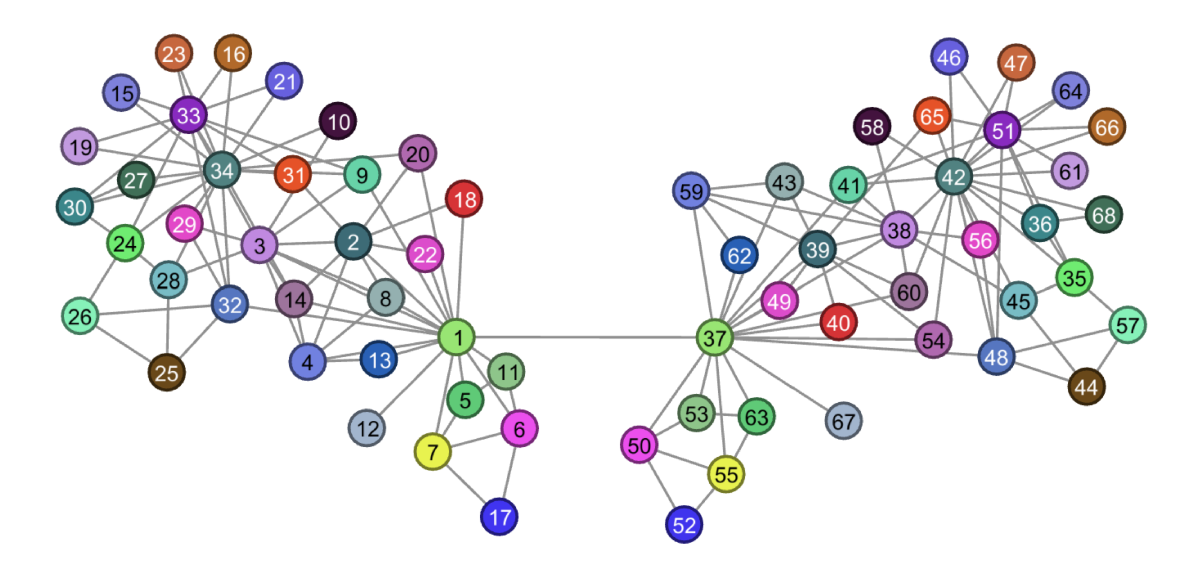}
\caption{Mirrored \textit{Zachary's Karate Club} network. Nodes and their mirrored counterparts are represented with identical colors to illustrate structural symmetry}
\label{fig:fig8}
\end{figure}

To ensure a fair comparison, the two networks were connected by an additional edge between the mirrored node pair $1$ and $37$. This modification was originally introduced because DeepWalk and node2vec cannot place nodes from different connected components in the same context. Figure~\ref{fig:fig8} illustrates the mirrored network, with corresponding node pairs highlighted in the same color. Nodes $1$ and $34$, along with their corresponding mirrored nodes ($37$ and $42$), represent the club instructor, Mr. Hi, and his administrator, John A. The network was constructed following a conflict between the two, which divided the club members into two factions, each centered around either Mr. Hi or John A. Consequently, nodes $1$ and $34$ occupy prominent central roles as leaders within their respective groups.

\begin{figure}[h]
\centering
\includegraphics[width=119mm]{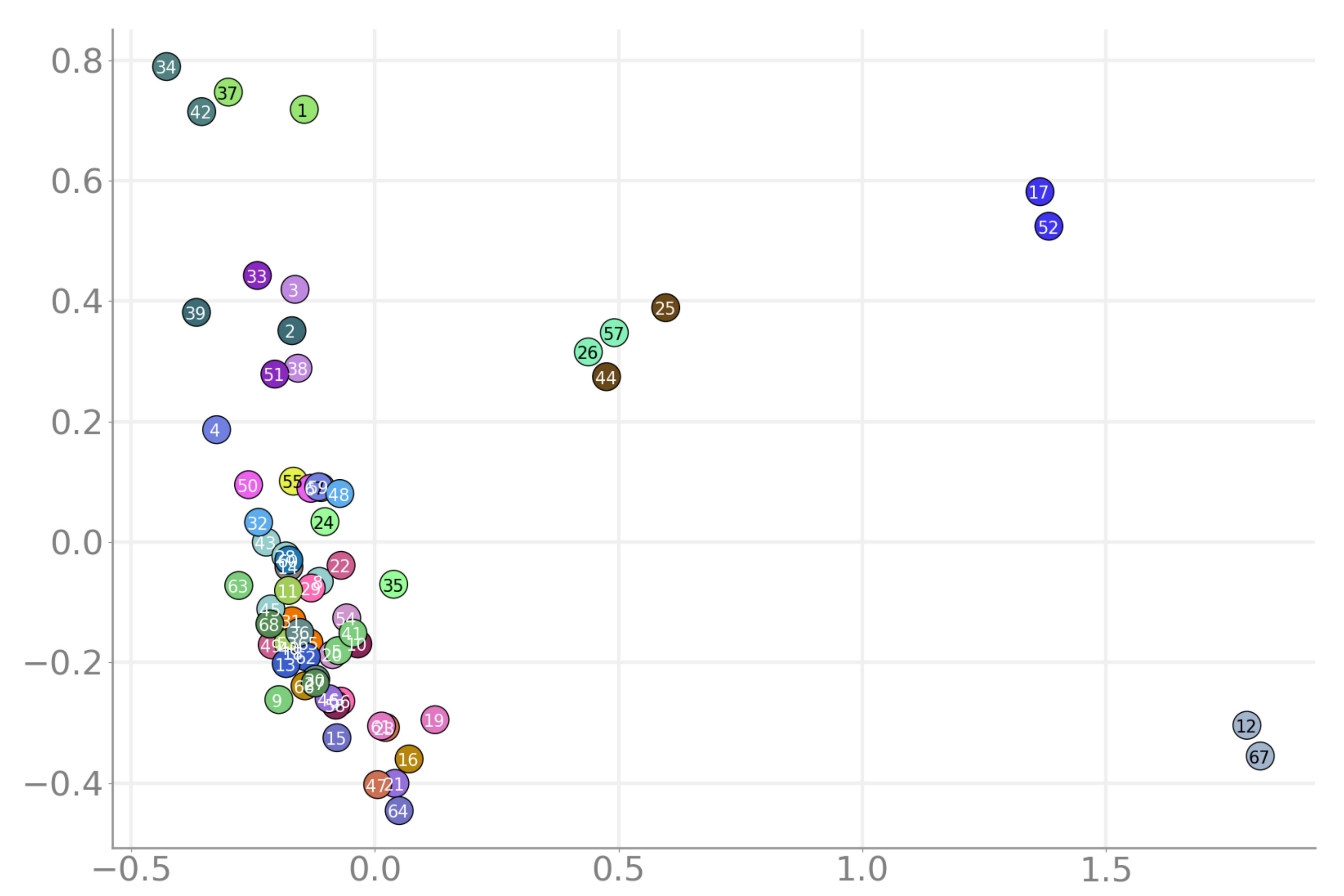}
\caption{ffstruc2vec node embedding vectors of the mirrored \textit{Zachary's Karate Club} network. Nodes and their mirrored counterparts share identical colors. ffstruc2vec effectively clusters structurally similar nodes, preserving the network’s structural properties in the embedding space}
\label{fig:figure_zachary_ffstruc2vec}
\end{figure}

\begin{figure}[h]
\centering
\includegraphics[width=119mm]{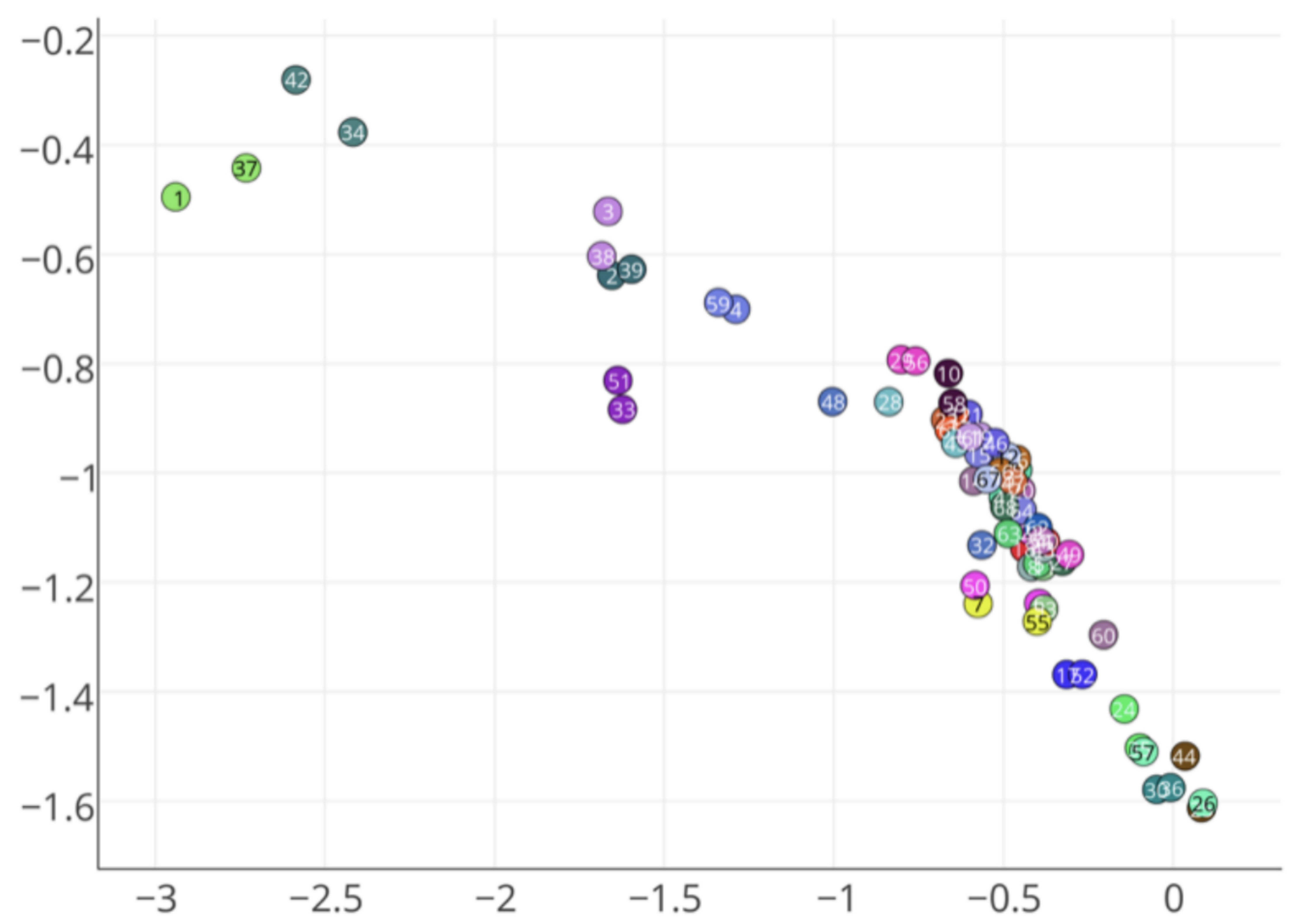}
\caption{struc2vec node embedding vectors of the mirrored \textit{Zachary's Karate Club} network. Nodes and their mirrored counterparts share identical colors. struc2vec fails to effectively separate several structurally distinctive nodes from the main cluster in the embedding space \citep{Ribeiro.2017}}
\label{fig:figure_zachary_struc2vec}
\end{figure}

The node embedding vectors of the mirrored Karate network, generated by \textbf{ffstruc2vec}, are visualized in two-dimensional space in \textbf{Figure~\ref{fig:figure_zachary_ffstruc2vec}}. Additionally, \textbf{Figure~\ref{fig:figure_zachary_struc2vec}} illustrates the node representation vectors produced by the \textbf{struc2vec} method, as presented in \citet{Ribeiro.2017}. While both methods capture the structural identity of nodes, it is evident that ffstruc2vec extracts additional structural patterns compared to struc2vec.

Specifically, ffstruc2vec clearly separates the node pair \textbf{(12, 67)} from the rest of the nodes, whereas struc2vec places this pair within the primary node group. Similarly, node pair \textbf{(17, 52)} is distinctly separated by ffstruc2vec, whereas struc2vec positions it at the periphery of the primary group. Furthermore, ffstruc2vec distinctly separates the node pairs \textbf{(25, 44)} and \textbf{(26, 57)} from the rest of the network, unlike struc2vec.

A structural analysis of these nodes, identified by ffstruc2vec as structurally outstanding, is provided below for further examination.

\begin{itemize}
	\item Nodes \textbf{12} and \textbf{67} possess a unique structural role within \textit{Zachary's Karate Club} network, as they are the only nodes with a degree of $1$.
	\item Nodes \textbf{17} and \textbf{52} possess a unique role as the only outstanding nodes associated with the central club instructor nodes \textbf{1} and \textbf{37}, positioned at the end of a small sub-cluster. This is evident from their assignment to these central nodes while being two hops away from them.
	\item Nodes \textbf{25}, \textbf{44}, \textbf{26}, and \textbf{57} possess a unique role as the only outstanding nodes associated with the central club administrator nodes \textbf{34} and \textbf{42}. This is evident from their assignment to these central nodes while being two hops away from them.
\end{itemize}

In summary, ffstruc2vec captures a broader range of structural roles than struc2vec, which aligns with its advantages discussed in Section~\ref{sec:Comparison_of_ffstruc2vec_with_struc2vec}.

Furthermore, ffstruc2vec and struc2vec successfully extracted multiple structural properties from the Karate network. Both methods were able to group mirrored node pairs—i.e., nodes of the same color—such that they remain close together in the latent space. Additionally, both methods identified nodes \textbf{1} and \textbf{34}, along with their corresponding mirrored nodes (\textbf{37} and \textbf{42}), as a separate cluster in the latent space. These nodes represent the central club instructor and the central club administrator. Moreover, ffstruc2vec and struc2vec identified nodes \textbf{2}, \textbf{3}, and \textbf{33}, along with their mirrored counterparts (\textbf{38}, \textbf{39}, and \textbf{51)}, as distinct from the rest of the network. These nodes also exhibit distinct structural identities, playing central roles within the network and possessing high node degrees.

\begin{figure}[h]
\centering
\includegraphics[width=119mm]{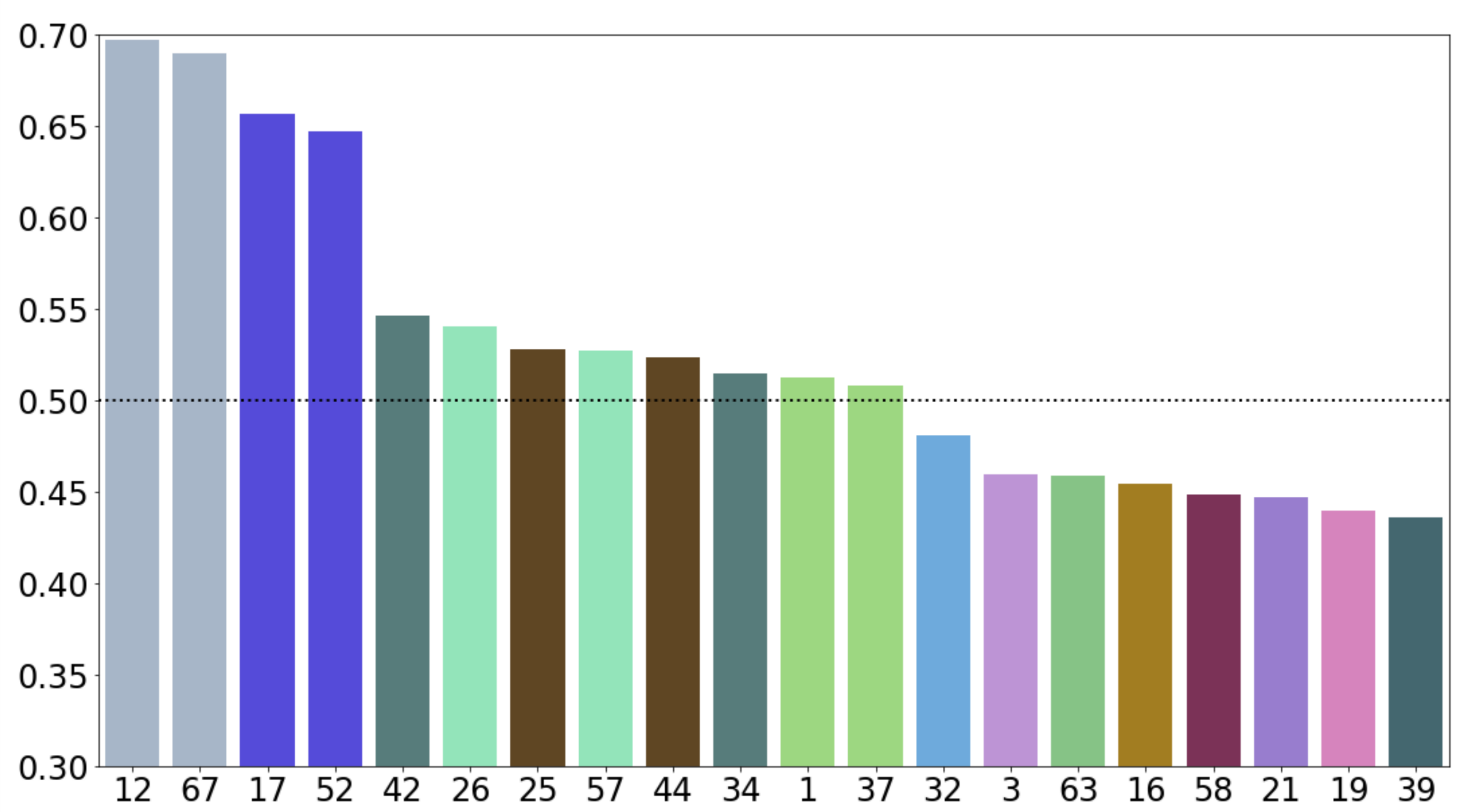}
\caption{Anomaly scores of the $20$ structurally most anomalous nodes in the mirrored Zachary's Karate Club network, computed using ffstruc2vec’s representation vectors. The y-axis denotes the anomaly scores, while the x-axis represents the corresponding nodes}
\label{fig:figure_zachary_outlier_scores}
\end{figure}

\textbf{Figure~\ref{fig:figure_zachary_outlier_scores}} presents the top 20 absolute anomaly scores for the nodes obtained by applying an \textbf{Isolation Forest} to the \textbf{ffstruc2vec} representation vectors. A higher score indicates a greater likelihood that a node is an outlier with respect to its structural embedding within the graph. Notably, due to their distinct roles in the mirrored \textit{Zachary's Karate Club} network, the nodes identified as structurally outstanding in the structural analysis above exhibit anomaly scores exceeding $0.5$.

\begin{figure}[h]
\centering
\includegraphics[width=119mm]{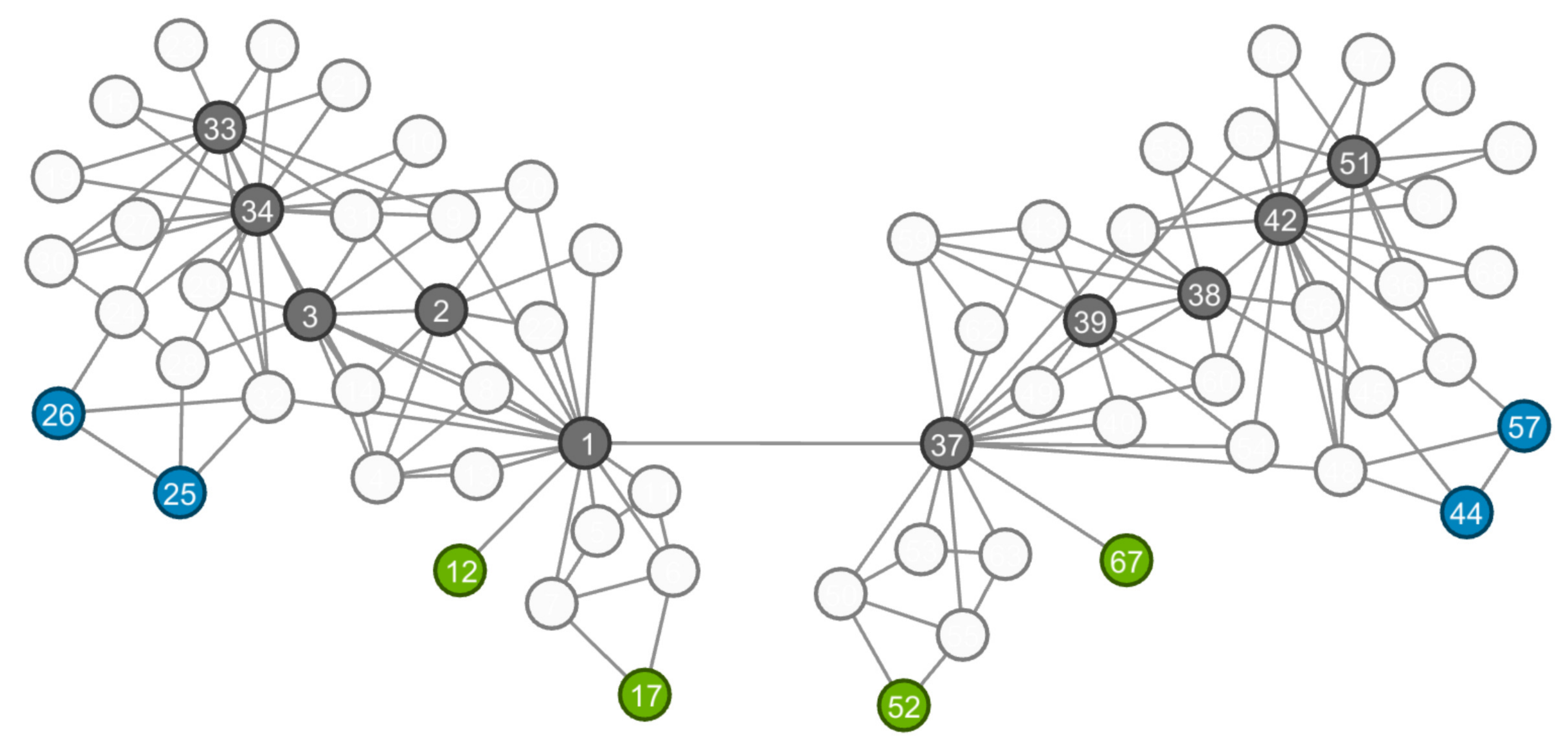}
\caption{Clustering of nodes in the mirrored \textit{Zachary's Karate Club} network using ffstruc2vec’s representation vectors. The clusters obtained using k-means group nodes with similar structural properties and are color-coded to reflect their structural similarity}\label{fig:figure_zachary_clustering}
\end{figure}

Applying a \textbf{k-means clustering} algorithm to the node embedding representations generated by \textbf{ffstruc2vec} to extract four clusters with similar structural properties results in the clusters shown in \textbf{Figure~\ref{fig:figure_zachary_clustering}}, each represented by a different color. It is visually evident that the clusters formed from these embeddings correspond to specific roles within the graph.

\begin{itemize}
    \item \textbf{White nodes} represent ordinary nodes that do not exhibit outstanding structural properties distinguishing them from the majority of the network.
    \item \textbf{Gray nodes} occupy certain central roles within the network.
    \item \textbf{Blue and green nodes} are separated from the other nodes due to the specific structural properties described above, which \textbf{ffstruc2vec} was able to extract more effectively than \textbf{struc2vec}.
\end{itemize}

\begin{figure}[h]
\centering
\includegraphics[width=119mm]{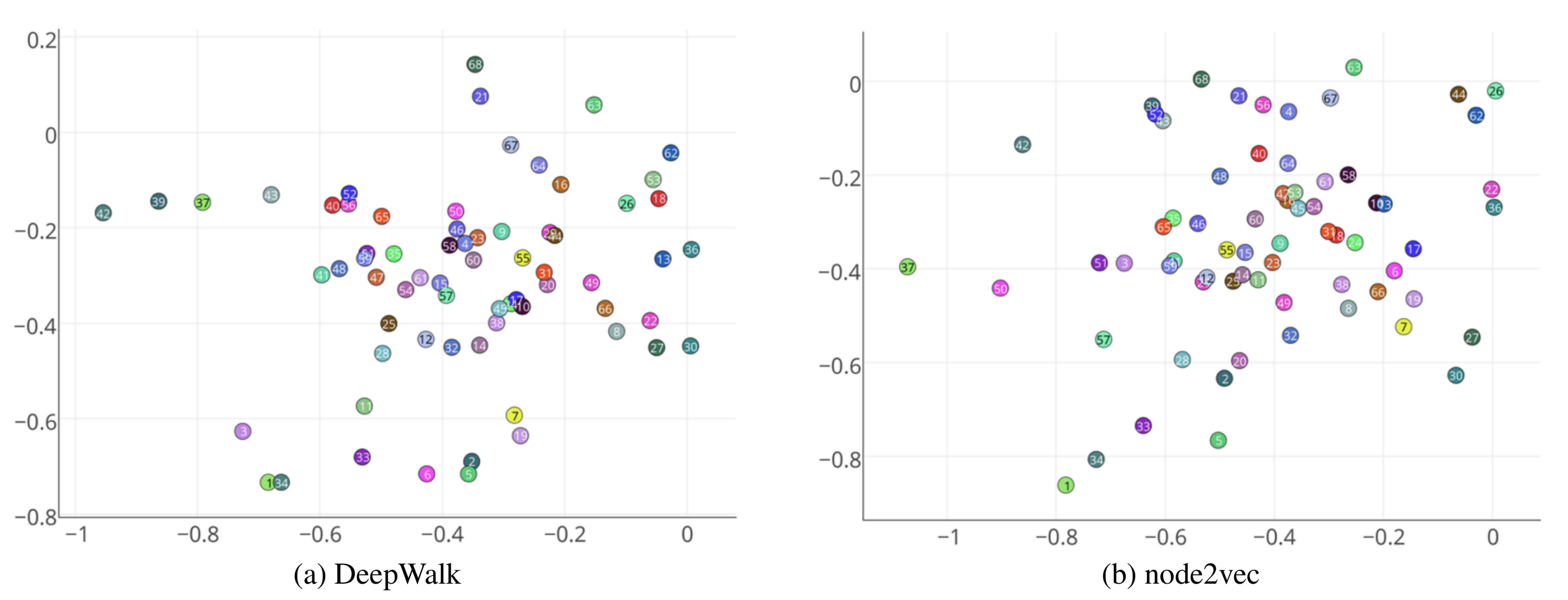}
\caption{Node embedding vectors of the mirrored \textit{Zachary's Karate Club} network, obtained using DeepWalk and node2vec. Nodes and their mirrored counterparts share identical colors. The embeddings produced by these methods fail to capture structural patterns \citep{Ribeiro.2017} }
\label{fig:figure12_zachary}
\end{figure}

A comparison of the node embeddings generated by \textbf{ffstruc2vec} and \textbf{struc2vec} with those of \textbf{DeepWalk} and \textbf{node2vec} reveals that the embeddings from \textbf{DeepWalk} and \textbf{node2vec} fail to capture these structural patterns and do not group structurally equivalent nodes in the latent space, as shown in \textbf{Figures \ref{fig:figure12_zachary}(a) and \ref{fig:figure12_zachary}(b)}.

\subsection{Barbell Graph}
\label{sec:Barbell_graph}

\begin{figure}[h]
\centering
\includegraphics[width=119mm]{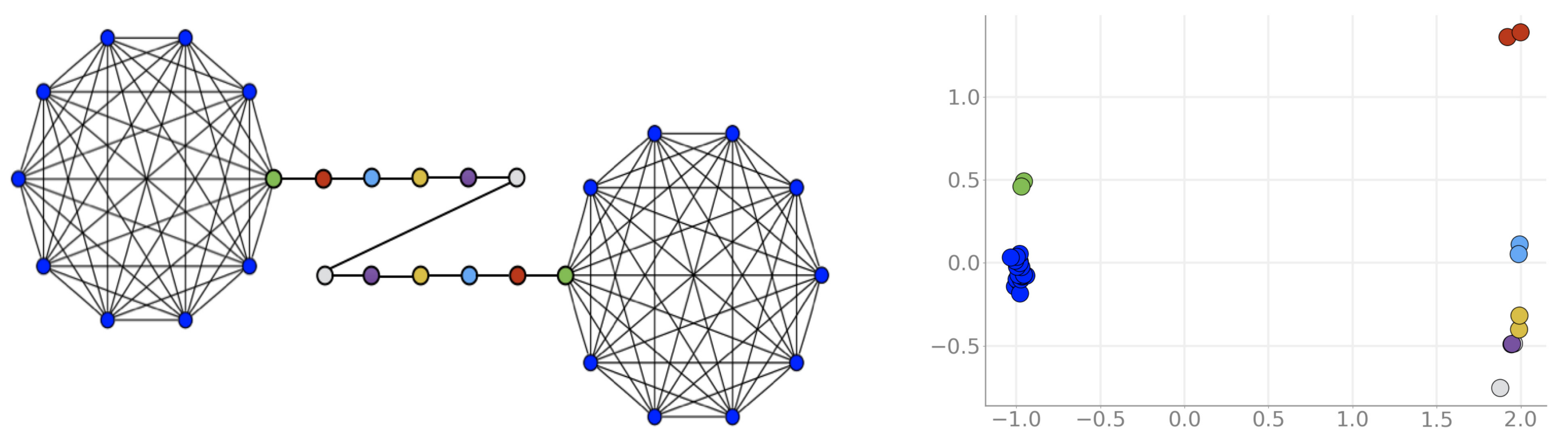}
\caption{Application of ffstruc2vec to a Barbell graph (left), demonstrating structural identity preservation in the embedding space (right). Nodes with the same structural identity in the Barbell graph share the same color. ffstruc2vec maps node representation vectors of structurally similar nodes to nearby positions in the embedding space}
\label{fig:figure_barbell_ffstruc2vec}
\end{figure}

\textbf{Figure~\ref{fig:figure_barbell_ffstruc2vec}} illustrates the application of the \textbf{ffstruc2vec} algorithm to a \textbf{Barbell graph}. Nodes with similar structural identities, as shown in the left image, are distinguished by uniform coloring. The \textbf{ffstruc2vec} algorithm effectively separates the embedding vectors of these nodes from the other nodes and positions nodes with similar structural identities closer to each other, as demonstrated in the right image.

The graph consists of two cliques, each comprising $10$ nodes, interconnected by a path of $10$ nodes. The nodes within the cliques that share identical structural identities are denoted in \textbf{blue}, while the nodes linking the cliques to the path are marked in \textbf{green}. \textbf{ffstruc2vec} positions the vectors of the \textbf{blue} nodes of the cliques closely together. In contrast, due to their additional connections to the path, the \textbf{green} nodes are positioned slightly apart from the blue nodes but remain relatively close, as expected.

Each node within the path has two neighbors and is colored \textbf{red}, \textbf{blue}, \textbf{yellow}, \textbf{purple}, or \textbf{white}, corresponding to its increasing distance from the cliques. As expected, \textbf{ffstruc2vec} places the nodes of the path far from the nodes of the cliques while preserving the relative order between them based on their distances to the cliques.

\begin{figure}[h]
\centering
\includegraphics[width=119mm]{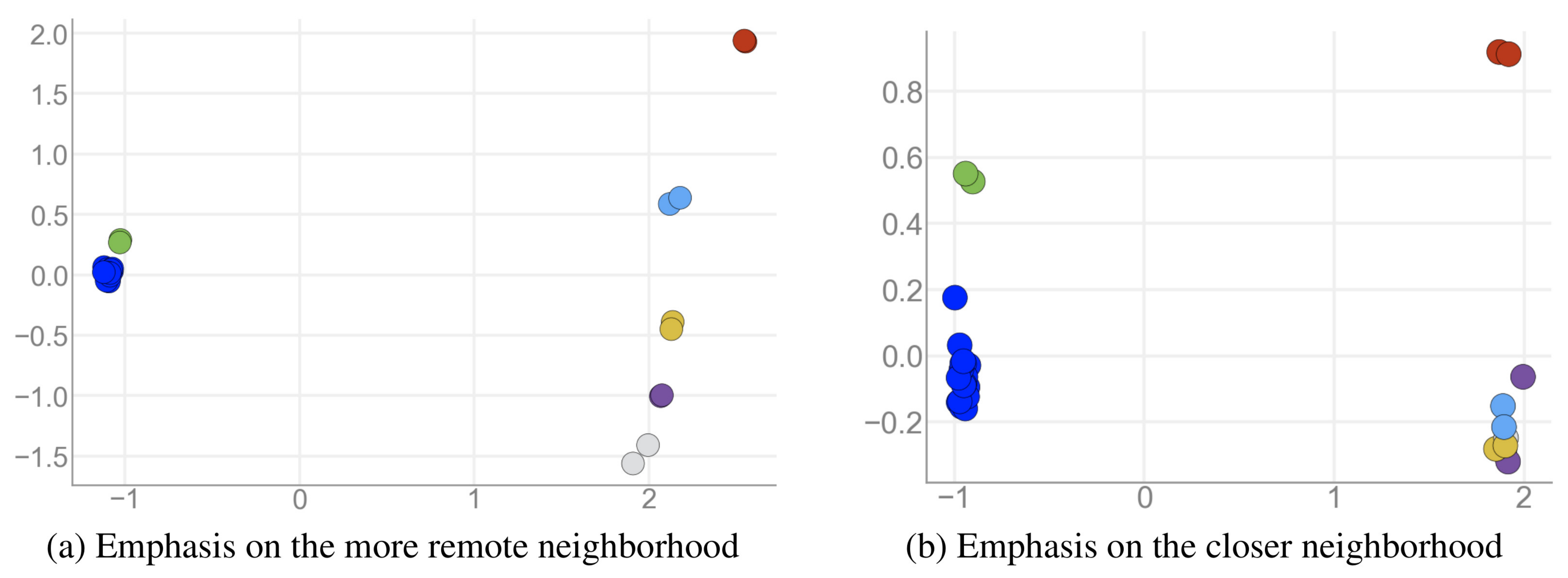}
\caption{Demonstration of ffstruc2vec’s flexibility in the Barbell graph by adjusting the weighting of k-hop neighborhood layers. In (a), higher weighting is assigned to distant neighborhoods, emphasizing long-range structural similarities. In (b), closer neighborhoods are prioritized, enhancing local structural differentiation}
\label{fig:figure_barbell_flexibility}
\end{figure}

\textbf{Figure~\ref{fig:figure_barbell_flexibility}} demonstrates \textbf{ffstruc2vec’s flexibility} in aligning with the demands of various downstream application tasks by emphasizing different types of structural identities. This adaptability is achieved by adjusting the weighting of $k$-hop neighborhood layers, which influences how structural similarities are captured in the latent space.

In \textbf{Figure~\ref{fig:figure_barbell_flexibility} (a)}, higher weighting is assigned to more distant $k$-hop neighborhood layers, leading to improved separation of nodes near the middle of the Barbell graph path, such as the \textbf{blue}, \textbf{yellow}, \textbf{purple}, and \textbf{white} nodes. This effect occurs because these nodes share a similar local neighborhood structure, meaning their distinction arises solely from the structural properties of their more distant neighborhoods. By prioritizing these higher-hop connections, ffstruc2vec enhances its ability to differentiate nodes based on structural roles beyond immediate proximity.

Conversely, in \textbf{Figure~\ref{fig:figure_barbell_flexibility} (b)}, higher weighting is applied to closer $k$-hop neighborhood layers. As a result, ffstruc2vec extracts similar structural identities for the \textbf{blue}, \textbf{yellow}, \textbf{purple}, and \textbf{white} nodes while simultaneously enabling better separation of the blue and green nodes. This shows how modifying the weighting scheme can shift the emphasis from long-range structural properties to local neighborhood similarities, highlighting ffstruc2vec’s adaptability in capturing different structural identities.

In contrast, \textbf{Figure~\ref{fig:figure_barbell_ffstruc2vec}} applies equal weighting to the first five $k$-hop neighborhoods when generating node representation vectors. This uniform weighting approach fails to highlight structural nuances as effectively as the adaptive weighting strategies shown in \textbf{Figure~\ref{fig:figure_barbell_flexibility}}. The comparison highlights the advantage of flexibly adjusting neighborhood influences, enabling ffstruc2vec to extract more meaningful representations by aligning with the requirements of the corresponding downstream application task.

\begin{figure}[h]
\centering
\includegraphics[width=119mm]{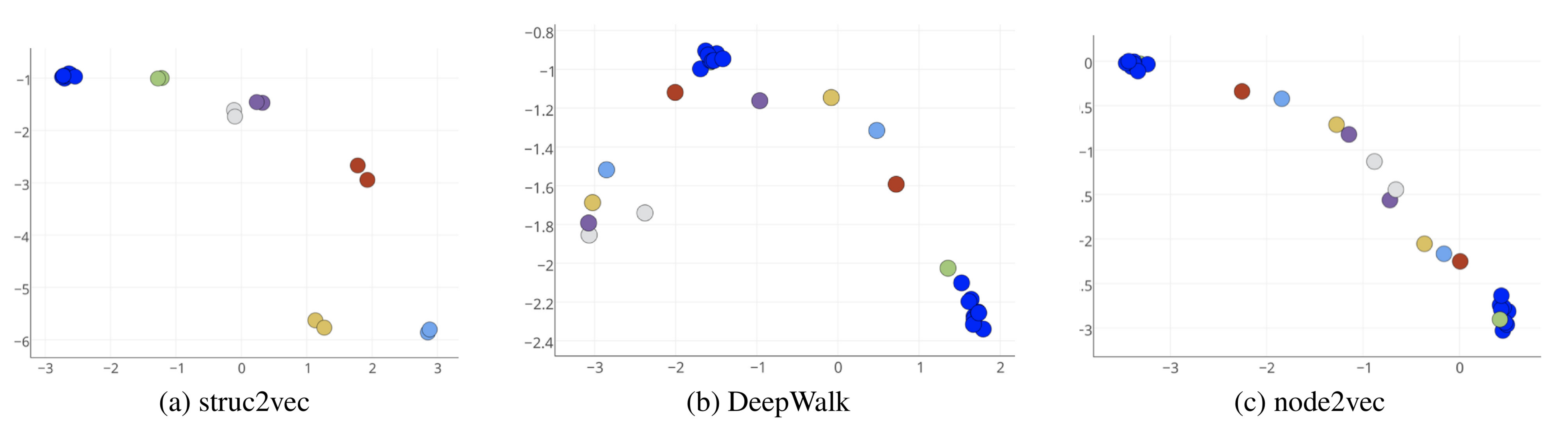}
\caption{Latent representations of nodes in the Barbell graph learned by three different node embedding frameworks \citep{Ribeiro.2017} }
\label{fig:figure_barbell_others}
\end{figure}

\textbf{Figure~\ref{fig:figure_barbell_others}} presents the latent representations for the nodes of the Barbell graph obtained using \textbf{struc2vec}, \textbf{DeepWalk}, and \textbf{node2vec}, as reported by \citet{Ribeiro.2017}. \textbf{Struc2vec} demonstrates structural preservation by grouping structurally equivalent nodes of the same color in the same vicinity in the latent space (Figure~\ref{fig:figure_barbell_others}(a)). However, it falls short compared to \textbf{ffstruc2vec} (Figure~\ref{fig:figure_barbell_ffstruc2vec}), as highlighted by the following discrepancies.

\begin{itemize}
    \item The nodes' structural identities in the Barbell graph's path are distinguished only by their distance to the two cliques. While \textbf{ffstruc2vec} maintains the relative ordering of red to white nodes, \textbf{struc2vec} fails to do so consistently in the embedding space.
    \item The nodes belonging to the two cliques in the Barbell graph have fundamentally different structural identities compared to the path nodes (e.g., $degree\geq 8$ vs. $degree=2$). In the embedding space, \textbf{ffstruc2vec} effectively separates these two groups, whereas \textbf{struc2vec} does not. For example, \textbf{struc2vec} positions the purple nodes significantly closer to the clique nodes than to the yellow nodes, even though the graph structure suggests that their structural identity is more aligned with that of the yellow nodes (see left image in Figure~\ref{fig:figure_barbell_ffstruc2vec}).
\end{itemize}

Section~\ref{sec:Comparison_of_ffstruc2vec_with_struc2vec} provides a detailed discussion of the limitations of \textbf{struc2vec} in extracting structural identities compared to \textbf{ffstruc2vec}.

For completeness, \textbf{Figure~\ref{fig:figure_barbell_others}} also includes the embeddings generated by \textbf{DeepWalk} (Figure~\ref{fig:figure_barbell_others}(b)) and \textbf{node2vec} (Figure~\ref{fig:figure_barbell_others}(c)), which were used as a baseline comparison in the \textbf{struc2vec} paper. Both fail to preserve the structural properties of the nodes, despite varying the parameters $p$ and $q$, which are designed in \textbf{node2vec} to control the balance between local and global exploration and can be adjusted to emphasize structural identities.

\subsection{Air-Traffic Network}
\label{sec:Air-traffic_network}

The node representation vectors generated by ffstruc2vec can serve as input for node classification tasks, provided that the labels for nodes correspond to the type of structural identities that ffstruc2vec can preserve. To evaluate its effectiveness, particularly in comparison to struc2vec, we use three benchmark datasets introduced in the struc2vec paper by \citet{Ribeiro.2017}. These datasets consist of unweighted, undirected air-traffic networks, where nodes represent airports and edges indicate the presence of commercial flights. Each airport is assigned one of four labels based on its activity level, measured by the number of flights or passengers. Specifically, the empirical activity distribution was divided into quartiles, with label $1$ assigned to the $25\%$ least active airports and label $4$ to the $25\%$ most active. All classes are of equal size and correspond to some type of structural role of the airport within the air-traffic network. The details of the three benchmark datasets are listed in Table~\ref{tab:table_Dataset_statistics}.

\begin{table}[h]
    \centering
    \caption{Overview of the three air-traffic network datasets used for evaluation. Each dataset is an unweighted, undirected graph where nodes represent airports and edges indicate commercial flight connections. The activity measurement is the basis for assigning activity-level labels to airports for the classification algorithm}
    \label{tab:table_Dataset_statistics}
    \begin{tabular}{lrrp{5cm}}
        \toprule
        Dataset & Nodes & Edges & Activity measurement \\
        \midrule
        Brazilian air-traffic network & 131 & 1,038 & Number of landings plus takeoffs \\
        American air-traffic network & 1,190 & 13,599 & Number of passengers that passed the airport \\
        European air-traffic network & 399 & 5,995 & Number of landings plus takeoffs \\
        \botrule
    \end{tabular}
\end{table}

Although a regression algorithm might yield better results, we follow the classification approach established in the struc2vec paper to ensure better comparability between the frameworks. To generate node embeddings for each air-traffic network, we applied ffstruc2vec, struc2vec, node2vec, and DeepWalk. The resulting embeddings were then used to train a supervised vector-based classification model. Instead of the grid search approach used in the struc2vec paper, we employed the Tree-structured Parzen Estimator (TPE) \citep{Bergstra.2011} to optimize hyperparameters, including the selection of the most suitable classification algorithm, thereby ensuring a fair benchmarking of all embedding methods. We considered node degree as the sole input feature, as it is a reasonable structural property for this task. Following the methodology of the struc2vec paper, we assessed performance using the accuracy score, given the balanced class distribution. Each experiment was repeated ten times with random training samples, using $80\%$ of the nodes for training. Moreover, we report the average performance across runs.

\begin{table}[h]
    \centering
    \caption{Classification accuracy of node embedding methods across three air-traffic network datasets. The best-performing method for each dataset is in bold}
    \label{tab:table_air_traffic_results}
    \begin{tabular}{lrrrr}
        \toprule
        Algorithm & Brazil & Europe & America & Average \\
        \midrule
        DeepWalk & 53.3\% & 48.1\% & 62.4\% & 54.6\% \\
        node2vec & 58.9\% & 50.8\% & 66.2\% & 58.6\% \\
        Node degree & 81.1\% & 57.0\% & 58.2\% & 65.4\% \\
        struc2vec & 75.9\% & \textbf{61.6}\% & 67.1\% & 68.2\% \\
        ffstruc2vec & \textbf{82.6}\% & \textbf{61.6}\% & \textbf{69.7}\% & \textbf{71.3}\% \\
        \botrule
    \end{tabular}
\end{table}

Table~\ref{tab:table_air_traffic_results} summarizes the classification performance of the examined methods across all air-traffic networks. DeepWalk yielded the lowest accuracy, as it primarily captures node proximity. For node2vec, the TPE algorithm favored larger values of $p$ and smaller values of $q$, indicating a preference for structural properties over proximity. Consequently, node2vec achieved a $4\%$ higher average accuracy than DeepWalk by emphasizing structural features. Incorporating pure node degree as an input feature for the classification algorithms led to an additional $6.8\%$ improvement in accuracy compared to node2vec.

Struc2vec, which explicitly considers the structural properties of node neighborhoods, further enhanced the average accuracy by $2.8\%$. The highest accuracy was achieved with ffstruc2vec, which outperformed struc2vec by $3.1\%$ due to its optimized flat framework and greater flexibility in capturing diverse structural identities.

\section{Conclusion}
\label{sec:Conclusion}

In this work, we introduced ffstruc2vec, a novel node embedding framework that effectively preserves structural identities by encoding a principled integration of multiple structural properties of nodes and their neighborhoods into an auxiliary flat similarity graph. By leveraging customizable comparison functions, even complex structural attributes can be seamlessly incorporated---enhancing the expressiveness of the learned embeddings.

A key strength of ffstruc2vec lies in its flexibility and interpretability. Most existing node embedding frameworks only focus on specific structural patterns or lack interpretability. In contrast, ffstruc2vec can be aligned to the specific requirements of downstream applications by optimizing weights---providing insight into the impact of individual structural characteristics on the given downstream application task. Our experimental evaluations on both supervised and unsupervised tasks confirm that ffstruc2vec outperforms existing methods in capturing structural identities.

To ensure scalability, ffstruc2vec achieves a time complexity of 
\[
\mathcal{O}(\max(|E|, |V| \cdot \log |V|))
\]
when incorporating $k$-hop neighborhoods for $k \in \{ 0,1 \}$. 

\paragraph{Future work.} We identify two key directions:
\begin{itemize}
  \item \textbf{Complexity analysis}: Optimize the current time complexity of $\mathcal{O}(\max(|E| \cdot |V|, |V| \cdot \log |V|))$ for structural patterns with $k \geq 2$ hops, e.g. through approximate neighborhood sampling.
  \item \textbf{Property aggregation}: Explore property aggregation strategies that balance the trade-off between representational power, flexibility, interpretability, and scalability.
\end{itemize}

\paragraph{Concluding Remarks.} 
By striking a balance between expressiveness, flexibility, interpretability, and scalability, ffstruc2vec establishes itself as a powerful framework for structural node embeddings.

\backmatter

\begin{appendices}

\renewcommand{\thefigure}{\arabic{figure}}
\setcounter{figure}{15} 

\renewcommand{\theequation}{\arabic{equation}}
\setcounter{equation}{7}

\section{Graph Indicators}
\label{sec:Graph_indicators}
ffstruc2vec utilizes graph indicators, as outlined in Section~\ref{sec:Measure_structural_similarity}, to extract structural patterns from a graph and encode these patterns in the node representation vectors for downstream application tasks. Specifically, ffstruc2vec allows the use of several graph indicators $I_i: V \to W(I_i)$ that provide metrics of structural properties of a node, for which a function 
\[
f_i: \mathcal{P}(W(I_i)) \times \mathcal{P}(W(I_i)) \to \mathbb{R},
\]
(see Equation~\ref{eq:simk_xy}) can be defined to compare the values of the graph indicator applied to two sets of nodes.

The selection of the appropriate graph indicators for a given downstream application task varies, depending on the specific structural patterns that need to be extracted. For example, in an air-traffic network (see Section~\ref{sec:Air-traffic_network}), the node degree (see Definition~\ref{def:node_degree}) may be the most suitable graph indicator to determine air traffic at airports, while in other tasks, a combination of more sophisticated graph indicators such as centrality measures may be more appropriate. Additionally, in other tasks, such as money laundering detection, more complex structures may be identified by graph indicators, such as graphlets surrounding a suspicious node in a transaction network. ffstruc2vec utilizes an optimized flat similarity graph, providing a high degree of flexibility in addressing the optimal integration of graph indicators. One approach, introduced in Section~\ref{sec:Measure_structural_similarity}, is using weighting factors that enable ffstruc2vec to prioritize relevant graph indicators for specific structural properties, addressing the requirements of specific downstream application tasks. The methods for learning and optimizing the weighting factors are discussed in Section~\ref{sec:Flexibility}. Evaluation of downstream application tasks utilizing different graph indicators within the ffstruc2vec framework is presented in Section~\ref{sec:Experimental_evaluation_and_benchmarking}. The following text presents some of the most relevant graph indicators, but the list can be expanded to extract additional structural patterns required for specific downstream application tasks.

\textbf{\textit{Node degree}}.
The degree of a node, as defined in Definition~\ref{def:node_degree}, is the number of edges connected to it. This graph indicator describes a structural property with low computational complexity. Depending on the specific application task, the degree of a node and the degree of its neighbors can provide information regarding the node's "role" and "importance" within the graph.

\textbf{\textit{Centrality measures}}.
While the node degree can provide information about the "role" and the "importance" of a node in a graph in certain application tasks, there are centrality measures that give a more sophisticated interpretation of these concepts. Various centrality measures exist, each interpreting the terms "role" and "importance" from different perspectives, such as the following.

\begin{itemize}
\item \textit{Closeness centrality} measures the proximity of a node to all other nodes in the graph.
\item \textit{Betweenness centrality} of a node is determined by the number of shortest paths between any two nodes that pass through it.
\item \textit{Eigenvector centrality}, as implemented in algorithms such as Google's PageRank \citep{Brin.1998} , determines a node's importance based on the importance of its neighboring nodes.
\item The \textit{core number} of a node quantifies its "coreness" within the graph. It is defined as the largest $k$ for which the node belongs to a $k$-core, where a $k$-core is a maximal subgraph in which all nodes have a degree of at least $k$.
\item For directed graphs, the "\textit{Authority}" and "\textit{Hub}" scores assess the importance of a node within the network. A node's authority measure increases with the number of important nodes pointing to it, while its hub measure increases with the number of important nodes it points to.
\end{itemize}

\textbf{\textit{Clustering Coefficient}}. 
The clustering coefficient of a node measures the extent to which its neighbors form a complete subgraph. It is defined as the ratio of the number of edges between the node’s neighbors to the total number of possible edges between them. A higher clustering coefficient indicates a more interconnected neighborhood.

\textbf{\textit{Graphlet Degree Vector (GDV)}}. 
The Graphlet Degree Vector (GDV) of a node is a numerical representation that characterizes the structural role of the node within a graph based on its local connectivity patterns. It is defined as a vector where each coordinate corresponds to the number of times the node participates in a specific orbit of a graphlet (see Definition~\ref{def:graphlets_and_orbits}). Graphlets are small, connected, non-isomorphic subgraphs, and orbits denote the distinct automorphism classes within a graphlet \citep{Milenkovi.2008}. 

\textbf{\textit{Anonymous walks}}. 
Anonymous walks (see Definition~\ref{def:Anonymous_Walk}) powerfully capture nodes' structural information. \citet{Micali.2016} demonstrated that anonymous walks capture characteristic structural patterns of graphs and allow for the exact reconstruction of a node's network proximity. Anonymous walks can be utilized in various ways to extract the structural properties of nodes.

\textbf{\textit{Random walks}}. 
Random walks can be utilized in various ways to extract the structural properties of nodes. For instance, after conducting random walks with various starting nodes within the graph, we define a new graph indicator, $I_i$, that counts the number of occurrences of a node in all the performed random walks. A higher number of occurrences indicates that the node is more structurally integrated within the graph and is more easily reachable from other nodes. Other examples of graph indicators $I_i$ based on random walks include counting the number of unique nodes in the random walks and counting the number of times the starting node appears in the anonymous walk.

\section{Integration of Node Proximity Properties and Node Features}
\label{sec:Integration_of_local_structural_properties}

The ffstruc2vec framework is primarily designed to capture the global structural properties of nodes. However, in many downstream applications, node proximity properties—such as community membership and shortest-path distances—also play a crucial role. ffstruc2vec allows for their integration by incorporating additional weighted summands into Equation~\ref{eq:sim_xy_final}. This extension enables the framework to incorporate community structures and shortest path information, as detailed in the following subsections.

Similarly, node features can be seamlessly integrated into ffstruc2vec using the same principle. Adding further weighted summands to Equation~\ref{eq:sim_xy_final} allows the framework to leverage additional node attributes while preserving its core structural embedding properties. This flexibility enhances ffstruc2vec's adaptability to various application scenarios, ensuring that structural and feature-based information can be effectively captured for downstream application tasks.

\subsection{Communities of Nodes}
\label{sec:Communities_of_nodes}

ffstruc2vec can be extended to incorporate community structures into node representation vectors, enhancing its ability to capture meaningful relationships for downstream application tasks. In many real-world graphs, nodes within the same community—defined as groups of nodes that are more densely connected to each other than to the rest of the network—often exhibit similar properties. Preserving community patterns in node embeddings can therefore improve their effectiveness in downstream application tasks. However, since there is no universally accepted definition of community structure, various approaches have been proposed for detecting optimal communities, each suited to different network types, application domains, and analytical objectives \citep{Yang.2010, Fortunato.2016, Khan.03.08.2017, Fang.2020}.

To incorporate community structures into node representations, an additional weighted summand can be introduced into Equation~\ref{eq:sim_xy_final}, using an indicator function $I_i: V \to W(I_i)$, where $I_i(v)$ represents the community assignment of node $v$. There are multiple ways to define the corresponding comparison function 
\[
f_i: \mathcal{P}(W(I_i)) \times \mathcal{P}(W(I_i)) \to \mathbb{R},
\]
One possible approach is to apply a vector-based community similarity.

For each set of nodes $I_i(N_k(x))$ and $I_i(N_k(y))$ (see Equation~\ref{eq:extended}), we construct a vector of length equal to the total number of communities. The $i$-th position of this vector represents the number of occurrences of the $i$-th community within the respective set of nodes. The comparison function $f_i$ can then be defined as the Euclidean distance between these vectors, quantifying the degree of co-membership of two nodes within the same community structure.

For $k=0$, $f_i$ assesses whether two nodes belong to the same community, effectively measuring node proximity. For $k\geq 1$, it evaluates the distribution of nodes in the $k$-hop neighborhoods of $x$ and $y$ across communities. Beyond simple community membership, these community-based vectors can also be interpreted as indicators of a node’s position within a cluster. For instance, if both $x$ and $y$ belong to the same cluster and are centrally located (meaning their $k$-hop neighborhoods also predominantly fall within the same cluster), they are assessed as highly similar. In contrast, their similarity decreases if one node is centrally located while the other is near the cluster boundary. Additionally, two nodes in different but adjacent clusters can still be considered more similar if their $k$-hop neighborhoods overlap, meaning they share a substantial number of common community memberships.

Beyond assessing node proximity, community structures can also serve as an indicator of another form of similarity. An alternative definition of $f_i$ measures the dissimilarity between the number of distinct communities in each node’s $k$-hop neighborhood. The diversity of community memberships within a node’s local neighborhood provides insight into the centrality of the node within its community and serves as an indicator of its proximity to other clusters.

\subsection{Shortest Path Between Nodes}
\label{sec:Shortest_path_between_nodes}
The shortest path between two nodes is the path with the minimum number of edges among all possible paths connecting them. In many real-world scenarios, nodes close to each other often exhibit similar properties. Consequently, preserving the shortest path distances in node representation vectors can enhance the effectiveness of downstream application tasks.

To incorporate this information, an additional weighted summand can be included in the structural similarity calculation (as outlined in Equation~\ref{eq:sim_xy_final}) without requiring a separate graph indicator $I_i$. This summand is defined by the comparison function

\[
f_i:V\times V\rightarrow\mathbb{N}, f_i(x,y)=lsp(x,y)
\]

\noindent
where $lsp:V\times V\rightarrow\mathbb{N}$ represents the length of the shortest path between two nodes $x,y\in V$, provided that a path exists between them. If no such path exists, its value should be set to a default that aligns with the requirements of the corresponding downstream application task.

Additionally, this summand can be extended to incorporate a comparative analysis of shortest path distances between neighboring nodes of $x$ and $y$. This is achieved by applying descriptive scalar aggregation functions (see Appendix~\ref{sec:Comparison_of_the_graph_indicators_of_node_groups}) to a pairwise comparison of the shortest paths between nodes in their respective $k$-hop neighborhoods.

\section{Functions for Comparing the Properties of Node Groups}
\label{sec:Comparison_of_the_graph_indicators_of_node_groups}

To determine the structural similarity $sim_k(x,y)$ between two nodes $x$ and $y$, as outlined in Equation~\ref{eq:simk_xy}, the properties of node groups, such as the node's graph indicator values $I_i: V \to W(I_i)$ that describe the structural properties of the nodes, as outlined in Appendix~\ref{sec:Graph_indicators}, are applied to the nodes of the $k$-hop neighborhoods of $x$ and $y$. The results are then compared utilizing the function 
\[
f_i: \mathcal{P}(W(I_i)) \times \mathcal{P}(W(I_i)) \to \mathbb{R},
\]
as outlined in Section~\ref{sec:Measure_structural_similarity}. The greater the output of the comparison function $f_i$, the more dissimilar the structural similarity between the two sets of nodes. The following discussion examines specific comparison functions $f_i$ that can be applied to the graph indicators in Appendix~\ref{sec:Graph_indicators}.

The graph indicators $I_i$, such as the node degree, centrality measures, clustering coefficient, number of different nodes in anonymous walks, number of times the starting node occurs in anonymous walks, or the number of occurrences of a node in random walks, as described in Appendix~\ref{sec:Graph_indicators}, have a value range of $W(I_i) \subset \mathbb{R}$ that characterizes the node's structural properties. Thus, we define functions $f_i : \mathcal{P}(\mathbb{R}) \times \mathcal{P}(\mathbb{R}) \to \mathbb{R}$ to measure the similarity between two sets of indicator outputs in $\mathbb{R}$.

\begin{itemize}
\item Let $A, B \in \mathcal{P}(W(I_i))$ be two input sets of real-valued graph indicator values.  
The function $f_i : \mathcal{P}(W(I_i)) \times \mathcal{P}(W(I_i)) \to \mathbb{R}$  
is defined as the Dynamic Time Warping (DTW) distance between the sorted sequences of $A$ and $B$, as proposed by \citet{Rakthanmanon.2013}:
\[
f_i(A, B) := \mathrm{DTW}(\mathrm{sort}(A), \mathrm{sort}(B))
\]
When applying DTW, various elementwise distance functions can be used to compare aligned elements. The choice of distance function depends on the characteristics of the data and the specific requirements of the downstream application task. However, since DTW was originally designed for time series analysis, it may not be well suited for extracting structural identities in many graph-based application tasks, as discussed in Appendix~\ref{sec:Advantages_of_ffstruc2vec_over_struc2vec}, point~(d).
\item We aggregate the values of a set using a descriptive scalar aggregation functions such as the mean, median, sum, minimum, maximum, variance, standard deviation, or interquartile range, or a weighted combination thereof and compare the results for the two sets, for example, using the difference or quotient. Furthermore, a weighted combination of aggregation functions can be optimized to fit a downstream application task.
\end{itemize}

The graph indicator GDV, as described in Appendix~\ref{sec:Graph_indicators},  
maps each node \( v \in V \) to a fixed-length vector over \( \mathbb{N} \), i.e.,  
\( \text{GDV} : V \to \mathbb{N}^d \),  
where \( d \) denotes the predefined number of considered graphlet orbits.  
Each entry in the vector represents the number of times the node appears in a specific orbit of a given graphlet.

To apply a function 
\[
f_i: \mathcal{P}(W(I_i)) \times \mathcal{P}(W(I_i)) \to \mathbb{R},
\]
to measure the similarity between the two sets of vectors of natural numbers, each set is first aggregated into a single representative vector using element-wise descriptive scalar aggregation functions (e.g., mean, median, maximum).  
The resulting aggregated vectors are then compared using a similarity or distance metric appropriate for the requirements of the downstream application task, such as:

\begin{itemize}
\item If absolute differences matter: \textit{Euclidean} or \textit{Manhattan distance}
\item If relative trends matter: \textit{Cosine similarity} or \textit{Pearson correlation}
\item If monotonic relationships matter: \textit{Spearman correlation}
\item If the data is categorical: \textit{Jaccard similarity} or \textit{Hamming distance}, e.g., communities of nodes (see Appendix~\ref{sec:Communities_of_nodes})
\end{itemize}

In addition to the method presented in Section~\ref{sec:Measure_structural_similarity},  
which incorporates structural properties as weighted terms into Equation~\ref{eq:sim_xy_final} by applying a comparison function to the values of graph indicators for two sets of nodes,  
similarity scores that directly compare two nodes can be incorporated in a similar manner.

Examples of such node-level comparison functions include the shortest path length (see Appendix~\ref{sec:Shortest_path_between_nodes}), the Katz index~\citep{Katz.1953}, SimRank~\citep{Jeh.2002}, MatchSim~\citep{Lin.2012}, and RoleSim~\citep{Jin.2011, Jin.2014}.

Let \( x, y \in V \) be the nodes whose structural similarity is to be measured.  
For \( k = 0 \), i.e., when considering only the nodes themselves, the similarity score between \( x \) and \( y \) is directly given by the respective comparison function.  
For \( k \geq 1 \), the similarity of their \( k \)-hop neighborhoods can be computed using descriptive scalar aggregation functions (e.g., mean, median, maximum) applied to all pairwise similarity values between nodes \( a \in N_k(x) \) and \( b \in N_k(y) \).

\section{Hyperparameters}
\label{sec:Hyperparameters}
The flexibility of ffstruc2vec in extracting various types of structural identities to suit different downstream application tasks is primarily determined by the following hyperparameters.

\begin{itemize}
\item
The integration of various graph indicators, as detailed in Appendix~\ref{sec:Graph_indicators}.
\item
Node proximity properties and node features can be incorporated, as described in Appendix~\ref{sec:Integration_of_local_structural_properties}, if relevant to the downstream application task.
\item
Various functions for comparing the structural properties of nodes can be applied, as outlined in Appendix~\ref{sec:Comparison_of_the_graph_indicators_of_node_groups}.
\item
The weighting of the utilized properties across different layers of the $k$-hop neighborhoods, as discussed in Section~\ref{sec:Measure_structural_similarity}.
\end{itemize}

The following list presents examples of additional hyperparameters in the framework that influence the types of structural identities extracted by ffstruc2vec. It is not exhaustive but intended as a representative selection of further hyperparameters.

\begin{itemize}
\item
In Section~\ref{sec:Generate_node_representation_vectors}, we generate node representation vectors using the Skip-Gram method. However, the proposed framework can also incorporate alternative techniques for learning latent representations from the given node sequences.
\item
The selection of the next node in a random walk step, as described in Section~\ref{sec:Performing_biased_random_walks}, depends on the edge weights of the similarity graph, which are derived from node similarities (see Equation~\ref{eq:sim_xy_final}). The edge weight $w_{xy}$ between nodes $x$ and $y$ in the similarity graph can be computed using various functions (see Section~\ref{sec:Construct_the_similarity-graph}). The choice of such a transformation function represents another hyperparameter of ffstruc2vec. For instance, a linear transformation can be applied, as shown in Equation~\ref{eq:trans_linear}.
\begin{equation}\label{eq:trans_linear}
w_{xy}=\dfrac{1}{sim(x,y)}
\end{equation}
A non-linear transformation can be applied using the function presented in Equation~\ref{eq:trans_non_linear}.
\begin{equation}\label{eq:trans_non_linear}
w_{xy}=wt^{-sim(x,y)}
\end{equation}
whereas the parameter $wt>1$ represents another hyperparameter. As $wt$ increases, the likelihood of selecting the node with the highest edge weight as the next step in the random walk increases, reducing bias. If $wt$ is set to Euler's number, the resulting distribution corresponds to the softmax function, which offers advantages for subsequent calculations due to its differentiability. As $wt$ continues to increase, the function approaches the $argmax$ function.

\end{itemize}

\section{Complexity Analysis}
\label{sec:Complexity_analysis}

We optimize the computational efficiency of the ffstruc2vec algorithm, achieving the following time complexity for the extraction of certain structural identities.
\[
\mathcal{O}(\max(|E|,|V|\cdot\log|V|))
\]
This is accomplished by refining the computation of structural similarities between node pairs, as outlined in Section~\ref{sec:Measure_structural_similarity}. Specifically, for each node $x\in V$, the similarity function $sim(x,y)$ is computed for only $\mathcal{O}(\log |V|)$ nodes $y\in V$, where $y\neq x$, that exhibit the highest similarity to $x$. This strategy constrains the selection of the next node in the random walk to the $\mathcal{O}(\log |V|)$ most similar nodes. Since these random walks aim to visit the most structurally similar nodes, this restriction preserves the quality of structural embeddings.

We restrict the optimization to the extraction of certain structural identities by applying the following two restrictions.
\begin{itemize}
\item We limit a node's structural identity computation to $k$-hop neighborhoods where $k \in \{ 0,1 \}$. This restriction ensures that structural properties are derived only from the node itself and its immediate neighborhood---regions that are typically the most informative for downstream application tasks.
\item ffstruc2vec allows the integration of arbitrary graph indicators and comparison functions. To achieve the time complexity of $\mathcal{O}(\max(|E|,|V|\cdot\log|V|))$, we constrain the graph indicators and comparison functions that are applied in Equation~\ref{eq:sim_xy_final} to those that can compute the structural similarity $sim_k(x,y)$ for all nodes $x,y\in V$ in $\mathcal{O}(\max(|E|,|V|\cdot\log|V|))$. This restriction still accommodates a broad range of graph indicators and corresponding comparison functions (see Appendices \ref{sec:Graph_indicators} and \ref{sec:Comparison_of_the_graph_indicators_of_node_groups}).
\end{itemize}

We demonstrate how the overall time complexity of $\mathcal{O}(\max(|E|,|V|\cdot\log|V|))$ can be achieved when utilizing the node degree (as defined in Definition~\ref{def:node_degree}) as a graph indicator and the difference between the mean node degrees of the nodes in $N_k(x)$ and $N_k(y)$ as a comparison function.

First, the degree of each node is computed in $\mathcal{O}(|V|+|E|)$ by initializing all degrees to zero and iterating over all edges, incrementing the degree of both adjacent nodes. The algorithm then applies $I_i$ to $N_k(x)$ for $k\leq k^*$, as defined in Equation~\ref{eq:sim_xy_final}. This means that the mean node degree of the nodes in $N_k(x)$ for $k\leq k^*$ must be computed for all $x\in V$.

For $k = 0$, the mean node degree of the nodes in $N_0(x)$ is already known for all $x \in V$, since $N_0(x) = \{ x \}$, meaning that the mean node degree in $N_0(x)$ equals the node degree of $x$.

For $k = 1$, the algorithm computes the average degree of the direct neighbors for each node in $\mathcal{O}(|V| + |E|)$ time. Specifically, for each node $x$, the algorithm iterates over all edges $e = (x, y) \in E$ and maintains two auxiliary values:
\begin{itemize}
    \item $m(x)$, representing the mean node degree of nodes in $N_1(x)$, and
    \item $n(x)$, representing the number of nodes in $N_1(x)$.
\end{itemize}
Both values are initialized to zero for all $x \in V$ and updated in constant time $\mathcal{O}(1)$ during each iteration, as stated in Equations~\ref{eq:update_m(x)} and~\ref{eq:update_n(x)}. Since there are $\mathcal{O}(|E|)$ iterations, each requiring $\mathcal{O}(1)$ time, the overall time complexity for computing the mean node degrees of nodes in $N_1(x)$ for all $x \in V$ remains $\mathcal{O}(|V| + |E|)$, assuming an efficient edge iteration mechanism, such as an adjacency list or edge list representation.

\begin{equation}\label{eq:update_m(x)}
	m(x):=\dfrac{m(x)\cdot n(x)+d_y}{n(x)+1}
\end{equation}

\begin{equation}\label{eq:update_n(x)}
	n(x):=n(x)+1
\end{equation}

To compute the structural similarity $sim(x,y)$, the algorithm then applies the comparison function $f_i$, as outlined in Equation~\ref{eq:sim_xy_final}. This function computes the difference between the mean values of two $k$-hop neighborhoods in constant time for each pair of nodes $(x,y)$, where $k\leq k^*$. To prevent a time complexity of $\mathcal{O}(|V|^2)$ when computing $sim(x,y)$ for all node pairs $x,y\in V$, the algorithm instead computes $sim(x,y)$ only for the $\mathcal{O}(\log |V|)$ nodes $y$ with the highest similarity to $x$ within a $k$-hop neighborhood for $k\leq k^*$.

To enable this selective computation, all nodes are sorted based on their degree statistics,  
which requires a time complexity of \( \mathcal{O}(|V| \cdot \log |V|) \) for each \( k \leq k^* \)~\citep{Cormen.2022}, resulting in \( k^* + 1 \) sorted lists \( l_k \).
These lists are then used to compute $sim_k(x,y)$ for the $\mathcal{O}(\log|V|)$ most similar nodes $y\in V$ to $x$ in $l_k$ for $k\leq k^*$. Instead of constructing a fully connected graph $G'$ as described in Section~\ref{sec:Construct_the_similarity-graph}, edges are created only between nodes $x,y\in V$ for which $sim_k(x,y)$ has been computed for at least one $k\leq k^*$. This results in a graph with $\mathcal{O}(|V|\cdot\log|V|)$ edges, where each node in $G'$ has at most $\mathcal{O}(\log|V|)$ edges. An edge weight in $G'$ is determined in constant time, resulting in a total time complexity of $\mathcal{O}(|V|\cdot\log|V|)$ for computing all edge weights.

To generate node representation vectors using $G'$, the algorithm first executes a constant number of biased random walks of fixed length for all $|V|$ nodes, as described in Section~\ref{sec:Performing_biased_random_walks}. Since each node in $G'$ has $\mathcal{O}(log|V|)$ neighbors, the upper bound on the time complexity for all random walks is $\mathcal{O}(|V|\cdot\log|V|)$. Subsequently, the algorithm applies Skip-gram to the sequences of nodes generated by these random walks (see Section~\ref{sec:Generate_node_representation_vectors}). The use of hierarchical softmax in the Skip-gram model reduces the time complexity to \( \mathcal{O}(|V| \cdot \log |V|) \)~\citep{Morin.2005}.

Applying a fixed number of ffstruc2vec iterations during the optimization process to identify suitable parameters---such as weighting factors---for a specific type of structural identities in a downstream application task (see Section~\ref{sec:Flexibility}) does not impact the time complexity.

Summarizing all described steps, ffstruc2vec achieves the following overall time complexity for the extraction of certain structural identities.
\[
\mathcal{O}(\max(|E|,|V|\cdot\log|V|))
\]

\section{Advantages of ffstruc2vec over struc2vec}
\label{sec:Advantages_of_ffstruc2vec_over_struc2vec}

This appendix expands on the advantages of ffstruc2vec over struc2vec, as outlined in enumerations (a) to (f) in Section~\ref{sec:Comparison_of_ffstruc2vec_with_struc2vec}.
\\
\\
\noindent
\textbf{(a) \& (b) Flexibility in Structural Identity Extraction through a Flat Similarity Graph}

\noindent
ffstruc2vec significantly enhances struc2vec by offering greater flexibility in extracting diverse structural identities tailored to different downstream application tasks (see Section~\ref{sec:Flexibility}. This is achieved by optimizing the weighting of multiple structural properties across different layers of the k-hop neighborhood, all of which are incorporated into a flat similarity graph (see Sections \ref{sec:Measure_structural_similarity} and \ref{sec:Construct_the_similarity-graph}).

In contrast, struc2vec constructs a randomized multilayer graph, where each layer corresponds to a $k$-hop neighborhood and edge weights are assigned incrementally. This approach prioritizes layers where nodes exhibit the highest degree of structural dissimilarity. However, this design does not always align with the specific requirements of downstream application tasks, as struc2vec may fail to focus on the most relevant layers. For instance, if a downstream application task relies predominantly on the structural properties of a specific $k$-hop neighborhood, struc2vec lacks a mechanism to emphasize this layer. Furthermore, struc2vec prioritizes layers where nodes exhibit higher structural dissimilarity. Since it applies Dynamic Time Warping (DTW) to measure similarity, and DTW is more likely to produce greater dissimilarities when processing a larger number of input nodes---which occurs more frequently in higher layers---the lower layers may not receive sufficient weight. This is problematic because lower layers often capture essential structural properties that are highly relevant for many downstream application tasks. Consequently, struc2vec may struggle to adequately preserve critical structural information, limiting its effectiveness in real-world applications.

Beyond layer selection, ffstruc2vec improves upon struc2vec by enabling the extraction of multiple weighted structural properties per layer (see Appendices \ref{sec:Graph_indicators} \& \ref{sec:Integration_of_local_structural_properties}), whereas struc2vec focuses on a single property, such as node degree. Moreover, ffstruc2vec integrates flexible comparisons of graph indicators across node groups (see Appendix~\ref{sec:Comparison_of_the_graph_indicators_of_node_groups}), allowing it to better capture relevant structural identities. As discussed in Appendix~\ref{sec:Hyperparameters}, this flexibility can be further extended through additional parameters.

The following examples illustrate how the advantages of ffstruc2vec enhance the creation of node embedding vectors.

One such example, presented in Section~\ref{sec:Zachary's_Karate_Club_unsupervised}, demonstrates how nodes $17$ and $52$ in the graph of mirrored \textit{Zachary’s Karate Club} (see Figure~\ref{fig:fig8}) stand out from all other nodes when considering node degree as a structural property. Although no other nodes have their respective karate club leader nodes (node $1$ or node $37$) appearing only in the $2$-hop neighborhood, struc2vec fails to distinguish the representation vectors of nodes $17$ and $52$ from the others (see Figure~\ref{fig:figure_zachary_struc2vec}). In contrast, ffstruc2vec successfully differentiates them (see Figure~\ref{fig:figure_zachary_ffstruc2vec}).

Another example, also based on the mirrored \textit{Zachary’s Karate Club} graph (see Figure~\ref{fig:fig8}), involves nodes $12$ and $67$, which have distinct structural properties. Although these nodes possess a degree of $1$, which is unique in the entire graph, and a unique $1$-hop neighborhood in \textit{Zachary’s Karate Club} graph, struc2vec fails to distinguish their node representation vectors from the others (see Figure~\ref{fig:figure_zachary_struc2vec}). Meanwhile, ffstruc2vec correctly identifies them as structurally distinct (see Figure~\ref{fig:figure_zachary_ffstruc2vec}).
\\
\\
\noindent
\textbf{(c) Interpretability and Explainability}

\noindent
An optimization process, which includes prioritizing and weighting graph indicators and $k$-hop neighborhoods, aligns ffstruc2vec to the specific requirements of a downstream application task.

The optimized values provide insights into the graph structures relevant to the specific application task, highlighting their impact, meaning, and relevance within the examined application scenarios (see Section~\ref{sec:Explainability_and_Interpretability}). In contrast, the struc2vec framework does not offer this capability.
\\
\\
\noindent
\textbf{(d) Limitations of Dynamic Time Warping (DTW) in struc2vec}

\noindent
struc2vec compares node degrees by applying Dynamic Time Warping (DTW), a technique traditionally used for time series analysis. The following example demonstrates a key limitation of DTW. 

\begin{figure}[h]
\centering
\includegraphics[width=119mm]{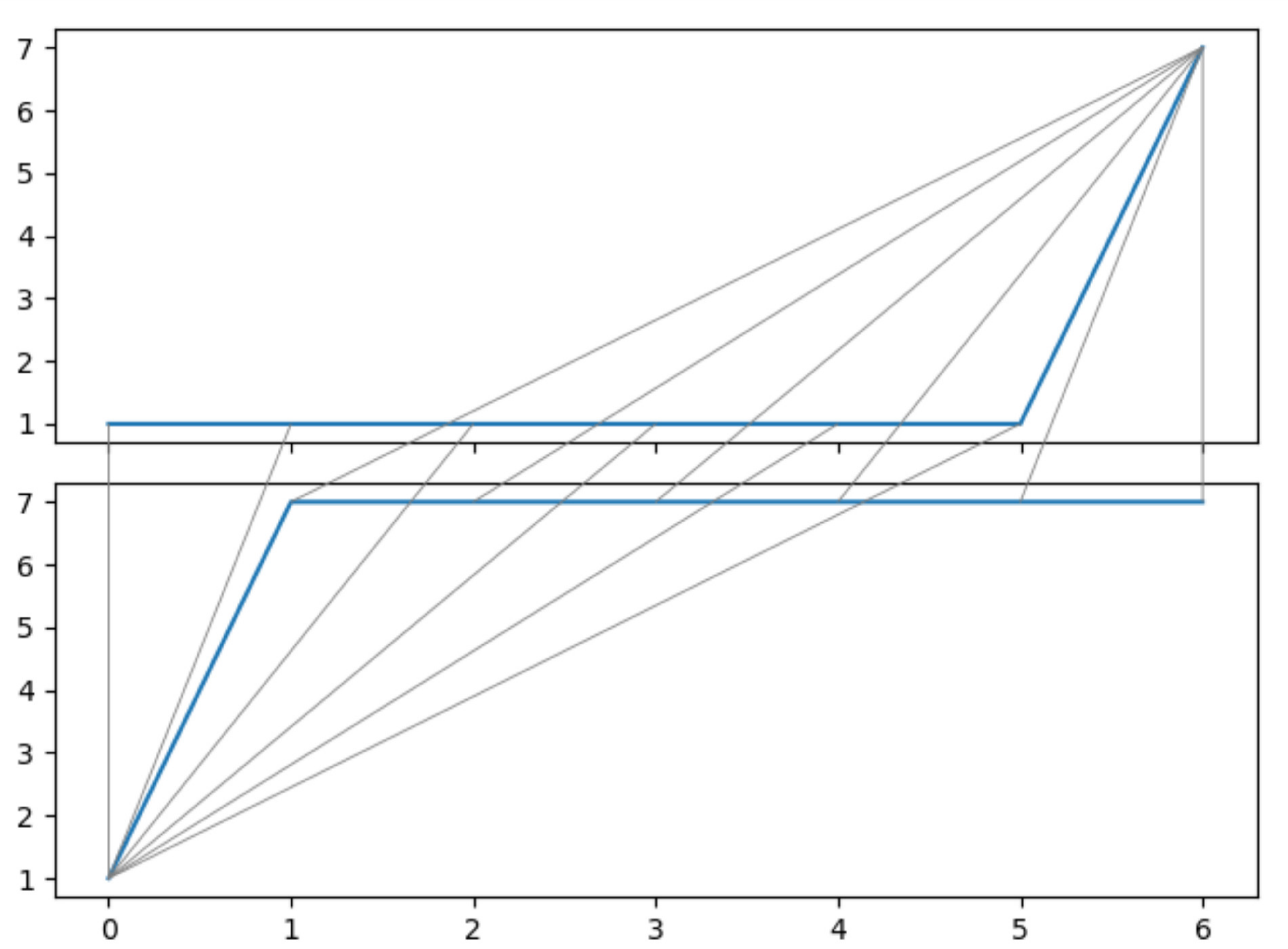}
\caption{Illustration of Dynamic Time Warping (DTW) applied to the node degree sequences 
$\{ 1, 1, 1, 1, 1, 1, 7 \}$ and $\{ 1, 7, 7, 7, 7, 7, 7 \}$. The grey lines indicate the element-by-element alignment, highlighting DTW’s limitation in distinguishing structurally different node groups}
	\label{fig:fig7}
\end{figure}

Consider a node group consisting of seven nodes with node degrees \(\{1, 1, 1, 1, 1, 1, 7\}\) compared to another group of seven nodes with node degrees \(\{1, 7, 7, 7, 7, 7, 7\}\). These groups should be distinguishable for most downstream application tasks. However, DTW calculates a difference of $0$ due to element-by-element comparisons of ones with ones and sevens with sevens, as symbolized by the grey lines in Figure~\ref{fig:fig7}. 

In most downstream application tasks, distinguishing these two node groups is essential since the second group is much more connected within the graph than the first group. A more suitable alternative to DTW in such cases is comparing the \textit{mean values} of node degrees between the two groups, which in this example are \(1.9\) and \(6.1\), respectively (see Appendix~\ref{sec:Comparison_of_the_graph_indicators_of_node_groups}).

Another drawback of DTW is that comparisons involving groups with a large number of nodes are more likely to produce higher dissimilarity values. Since struc2vec prioritizes layers where structural properties are highly dissimilar, it often struggles because structural patterns in the direct neighborhood are more crucial in many real-world applications, yet the direct neighborhood often contains fewer nodes compared to farther $k$-hop neighborhoods.

For elementwise comparison, struc2vec employs the distance measure function defined in Equation~\ref{eq:distance}, which amplifies differences in small node degrees (e.g., between $1$ and $2$) while diminishing differences in large node degrees (e.g., between $101$ and $102$). This property can be beneficial for certain applications when measuring node degree distances. However, in some application tasks, an absolute distance measure may be more suitable than a relative one.

For instance, in financial transaction networks used for fraud detection, where nodes represent bank accounts and edges represent transactions, a relative distance measure may treat a sudden increase from $1,000$ to $2,000$ transactions per day in a high-volume account as equally concerning as an increase from $10$ to $20$ in a low-volume account. To avoid downplaying fraud signals in high-activity accounts, an absolute distance measure may be preferable in such downstream application tasks.

\begin{equation}\label{eq:distance}
d(a,b)=\dfrac{\max(a,b)}{\min(a,b)}-1
\end{equation}
\\
\\
\noindent
\textbf{(e) Scalability}

\noindent The ffstruc2vec framework's scalability is effective for large graphs, such as those commonly found in social networks, which may contain billions of nodes and edges. The time complexity of the ffstruc2vec algorithm can be optimized to 

\[
\mathcal{O}(\max(|E|, |V| \cdot \log|V|))
\]
\noindent
for the extraction of certain structural identities, as shown in Section~\ref{sec:Scalability}, whereas the struc2vec algorithm exhibits a time complexity of 

\[
\mathcal{O}(|V|^3).
\]
\\
\\
\noindent
\textbf{(f) Improved Extraction of Structural Identities in Downstream Applications}
\noindent

The improved extraction of structural identities using ffstruc2vec is demonstrated in practical unsupervised downstream application tasks (see Sections \ref{sec:Zachary's_Karate_Club_unsupervised} and \ref{sec:Barbell_graph}) as well as in supervised downstream application tasks (see Section~\ref{sec:Air-traffic_network}), where ffstruc2vec significantly outperforms struc2vec.

\end{appendices}

\bmhead{Funding}
The authors did not receive support from any organization for the submitted work.

\bmhead{Competing Interests}
The authors have no relevant financial or non-financial interests to disclose.

\bmhead{Code availability}
The source code of ffstruc2vec is available online at 
\href{https://github.com/node-embedding/ffstruc2vec}{https://github.com/node-embedding/ffstruc2vec}.

\bmhead{Third-Party Content Permissions}
\hspace{1pt} \\[-0.4\baselineskip]

\noindent\textbf{Figure~\ref{fig:fig3}} is reproduced from: \\
Pržulj, N. (2007). \textit{Biological network comparison using graphlet degree distribution}. \textit{Bioinformatics}, 23(2), e177--e183. © Oxford University Press. Reproduced with permission (RightsLink License No. 5994890766474).

\medskip

\noindent\textbf{Figures~\ref{fig:figure_zachary_struc2vec}, \ref{fig:figure12_zachary}, and \ref{fig:figure_barbell_others}} are reproduced from: \\
Ribeiro, L. F. R., Saverese, P. H. P., \& Figueiredo, D. R. (2017). \textit{struc2vec: Learning node representations from structural identity}. In \textit{Proceedings of the 23rd ACM SIGKDD Conference on Knowledge Discovery and Data Mining (KDD '17)}. © ACM. Reproduced with permission (CCC License No. 1591085-1).


\end{document}